\documentclass[nonblindrev]{informs3} % current default for manuscript submission

\usepackage{balance}
\usepackage{amsmath}
\usepackage[tight,footnotesize]{subfigure}
\usepackage{graphicx}
\usepackage{bm}
\usepackage{algorithm}
\usepackage[noend]{algorithmic}
\usepackage{multirow}
\usepackage{diagbox}
\usepackage{adjustbox}
\usepackage{booktabs}
\usepackage{caption}
\usepackage{amsfonts}
\usepackage{siunitx}

\OneAndAHalfSpacedXII % current default line spacing
\usepackage{natbib}
 \bibpunct[, ]{(}{)}{,}{a}{}{,}%
 \def\bibfont{\small}%
\TheoremsNumberedThrough     % Preferred (Theorem 1, Lemma 1, Theorem 2)
\EquationsNumberedThrough    % Default: (1), (2), ...
\MANUSCRIPTNO{2015} % When the article is logged in and DOI assigned to it,
\begin{document}
%%%%%%%%%%%%%%%%

%\setcounter{page}{1}
%\MONTH{May}% (month or a similar seasonal id)
%\YEAR{2015}% e.g., 2005
%\SHORTYEAR{15}% shortened year (two-digit)
%\DOI{10.1287/xxxx.0000.0000}%

\RUNAUTHOR{Tang et al.}

\RUNTITLE{Group Equality in Adaptive Submodular Maximization}

% \TITLE{Bundling Infformation Goods of Decreasing Value}
\TITLE{Group Equality in Adaptive Submodular Maximization}

\ARTICLEAUTHORS{%
\AUTHOR{Shaojie Tang}
\AFF{Naveen Jindal School of Management, The University of Texas at Dallas}
\AUTHOR{Jing Yuan}
\AFF{Department of Computer Science and Engineering, The University of North Texas}
} % end of the block

\ABSTRACT{In this paper, we study the classic submodular maximization problem subject to a group equality constraint under both non-adaptive and adaptive settings. It has been shown that the utility function of many machine learning applications, including data summarization, influence maximization in social networks, and personalized recommendation,  satisfies the property of submodularity. Hence, maximizing a submodular function subject to various constraints can be found at the heart of many of those applications.
On a high level, submodular maximization aims to select a group of most representative items (e.g., data points). However, the design of most existing algorithms does not incorporate the fairness constraint, leading to under- or over-representation of some particular groups. This motivates us to study the submodular maximization problem with group equality, where we aim to select a group of items to maximize a (possibly non-monotone)  submodular utility function subject to a group equality constraint. To this end, we develop the first constant-factor approximation algorithm for this problem. The design of our algorithm is robust enough to be extended to solving the submodular maximization problem under a more complicated adaptive setting. Moreover, we further extend our study to incorporating a global cardinality constraint and other fairness notations.}

%\KEYWORDS{approximation algorithm; team formation; cover decomposition}

\maketitle
\section{Introduction}
Submodular maximization is  a fundamental discrete optimization problem which can be found at the heart of many machine learning and artificial intelligence applications. The property of submodularity, which captures the notion of diminishing returns, naturally occurs in a variety of real-world settings. To name a few, feature selection in machine learning \citep{das2008algorithms}, exemplar-based clustering \citep{dueck2007non}, active learning \citep{golovin2011adaptive}, influence maximization in social networks \citep{tang2020influence}, recommender system \citep{el2011beyond}, and data
summarization \citep{sipos2012temporal}.  This has made the design of effective and efficient algorithms for maximizing submodular functions increasingly important. Towards this end, extensive research has been conducted on developing good algorithms subject to a wide range of practical constraints, including cardinality, matroid, or knapsack-type restrictions. In this paper, we are interested in solving the classic submodular maximization problem subject to group fairness constraints. Given that there does not exist an universal metric of group fairness, we adopt the notation of \emph{group equality}, where we seek a \emph{balanced} solution across multiple groups. Formally, the input of our problem is a set $V$ of items (e.g., people). We partition $V$ into $m$ groups: $V_1, V_2, \cdots, V_m$, each group represents those items sharing the same attribute (e.g., race). We say that a set $S\subseteq V$ satisfies group equality if for all $i, j\in[m]$, we have $|S \cap V_i|-|S \cap V_j| \leq \alpha$, where $\alpha\in \mathbb{Z}_{\ge 0}$ is a group equality constraint.  Group equality captures the straightforward goal of balancing the number of items selected from each group.  Intuitively, one can adjust the degree of group equality through choosing an appropriate $\alpha$. For example, $\alpha=0$ leads to the highest degree of group equality because in this case, every feasible solution must contain the same number of items from each group; at the other end of the spectrum, if we set $\alpha=n$, then there is no group equality constraints. We next provide some relevant examples that adopts group equality. %One example is \emph{equal representation}  in the Senate \citep{enwiki:1115074261}. That is, the Senate of the United States shall be composed of the same number of Senators from each State regardless of their size/population.
One example is about the fairness in proposal solicitations. For many proposal solicitations from NSF (such as the recently announced ExpandAI program \citep{nsf}) and other agencies, each organization is requested to submit the same number of proposals, regardless of their sizes,  to those programs. The other example is about the fairness in the design of  hierarchical recommender systems for food delivery industry (e.g., Uber Eats) and video-on-demand services (e.g., Netflix).  For example, in Uber Eats the user is often provided with multiple groups of recommendations such as ``Chinese Food'' or ``Thai Food''. Each group of restaurants is displayed as carousels that allow the consumer to horizontally scroll between different restaurants. Note that every carousel contains roughly the same number of restaurants regardless of the type of the cuisine.
%Group equality can also be found in athletics, for example, every country's national soccer team must consist of the same number of players regardless of their population.

\textbf{Additional notes on group fairness.} Group fairness can be conceptualized in various ways, but typically falls into two primary categories as outlined by \citep{bronfenbrenner1973equality}: \emph{equality-fairness} and \emph{equity-fairness}. Equality-fairness involves offering the same resources or opportunities to every group without considering specific group attributes. On the other hand, equity-fairness acknowledges the unique situations of each group, ensuring they receive the exact resources necessary to achieve similar outcomes. While our primary focus leans towards equality-fairness - selecting an approximately equal number of items from each group irrespective of their sizes - we also address equity-fairness in an extension, ensuring that the number of items selected from each group $V_i$ is in a predefined range of $[\underline{\gamma_i}, \overline{\gamma_i}]$.

\subsection{Our Contributions}
Next, we outline the key contributions of this paper. An overview of our findings is provided in Table \ref{rrr}.
\begin{itemize}
\item We first study the problem of maximizing a non-monotone submodular function subject to  group equality constraints under the non-adaptive setting. It has been shown that many applications have non-monotone objectives, that is, adding an item to an existing solution may decrease its utility. This makes it challenging to design efficient algorithms with provable guarantees of performance. To tackle this challenge in the presence of group equality constraints, we first convert our problem to a carefully designed matroid constrained submodular maximization problem at the cost of losing a constant factor in the approximation ratio, then use a random greedy algorithm solve this new problem to obtain an approximate solution. Unfortunately, this solution may not be feasible to our original problem, to this end, we add some additional items to this solution to make it feasible. We show that the final solution achieves a constant approximation ratio for our original problem.
\item Then we study our problem in a more complicated adaptive setting. Following the framework of adaptive submodular maximization \citep{golovin2011adaptive}, our goal is to maximize a stochastic submodular objective, whose value is dependent on both the identities of selected items as well as their states. The biggest challenge under the adaptive setting is that the realized state of each item is unknown initially, one must select an item before observing its realization. Hence, a solution under the adaptive setting can be characterized as a \emph{policy} that maps the current observation to the next selected item. One important application of this framework is experimental design, where the input is a set of tests, the state of a test is the outcome from that test, then our goal is to \emph{adaptively} select (e.g., conduct) a group of tests to reach the most accurate conclusion about a patient. For this adaptive variant, we develop a policy that guarantees a constant factor approximation to the \emph{best adaptive} policy subject to the group equality constraint. To the best of our knowledge, we are the first to study the submodular optimization problem with group equality under the adaptive setting.
\item Next, we add a global cardinality constraint to our basic model. Formally, under this extended variant, we say that a solution $S$ is feasible if $|S \cap V_i|-|S \cap V_j| \leq \alpha, \forall i, j\in[m]$ and $|S|\leq c$ for some cardinality constraint $c$. We develop a constant-factor approximation algorithm for this variant under the non-adaptive setting.
\item At last, we discuss another frequently employed fairness notation referred to as ``equity-fairness constraints'', which ensures that the selection of items from each group falls within a predefined range. Specifically, we say that a solution $S$ is feasible if $|S \cap V_i| \in [\underline{\gamma_i}, \overline{\gamma_i}], \forall i \in[m]$ and $|S|\leq c$.  We develop approximation algorithms for this variant under the adaptive setting.
\end{itemize}

\begin{table}[t]
\begin{center}
\begin{tabular}{ |c|c|c|c| }
\hline
Setting & Region &Approximation ratio  & Section \\
\hline
  \multirow{3}{*}{non-adaptive}& $k_{\min}> 1$ & $0.045^*$ & \ref{sec:non-adaptive-general} \\
 & $k_{\min}= 0$ & $\frac{1}{e}-o(1)$  & \ref{sec:non-adaptive-special}; \ref{sec:non-adaptive-special1}\\
 &  $k_{\min}= 1$ & $(1/2)(\frac{1}{e}-o(1))$ & \ref{sec:non-adaptive-special}; \ref{sec:non-adaptive-special2}\\
\hline
non-adaptive (monotone) & all & $1-1/e$ & \ref{sec:non-adaptive-monotone}\\
\hline
%\citep{hanicml}& $\frac{1}{k+2\sqrt{k+1}+2}$ & $k$-system constraint & Yes \\
%\citep{tang2021beyond} & $\frac{1}{e}$ &cardinality constraint & No \\
  \multirow{4}{*}{adaptive}& $k_{\min}> 1$  & $1/24$ & \multirow{1}{*}{\ref{sec:adaptive-general}} \\
  &$k_{\min}= 0$ & $1/6$&\ref{sec:adaptive-special} \\
  &$k_{\min}= 1$ and $\alpha=0$& $1/12$&\ref{sec:adaptive-special1}\\
  &$k_{\min}= 1$ and $\alpha\geq 1$& $1/10$&\ref{sec:adaptive-special2}\\
%this work & $\frac{1}{2k+4}$ &  $k$-system constraint & No \\
 \hline
 %\citep{hanicml}& $\frac{1}{k+2\sqrt{k+1}+2}$ & $k$-system constraint & Yes \\
%\citep{tang2021beyond} & $\frac{1}{e}$ &cardinality constraint & No \\
adaptive (monotone)& all  & $1/2$ &\ref{sec:adaptive-monotone}\\
 \hline
extended non-adaptive&  all &  $(1/8)(\frac{1}{e}-o(1))$ &\ref{sec:non-adaptive-extended}; \ref{sec:extended-speical}\\
\hline
extended non-adaptive (monotone)& all &  $1-1/e$ &\ref{sec:extended-monotone}\\
\hline
adaptive with equity-fairness & all &  $(1-\max_{i\in[m]} \underline{\gamma_i}/k_i)/6$&\ref{sec:extended-equity}\\
\hline
\end{tabular}
\caption{Summary of our results. Here $k_{\min}=\min_{i\in[m]}|V_i|$ denotes the size of the smallest group.\\  $^*$ This ratio can be improved to $(1/8)(\frac{1}{e}-o(1))$ using the method developed for the extended non-adaptive model. It is important to note, however, that our proposed solution for the extended model has a drawback - it is not adaptable to the adaptive framework. Further information and elaboration on this matter can be found in Remark 2.}
\label{rrr}
\end{center}
\end{table}

\subsection{Additional Related Works}
There is an extensive literature on the development of fair algorithms for various applications, such as classification \citep{zafar2017fairness}, influence maximization \citep{tsang2019group}, bandit learning \citep{joseph2016fairness}, voting \citep{celis2018multiwinner}, and data summarization \citep{celis2018fair}.  Recently, \cite{el2020fairness} addressed fairness in the context of streaming submodular maximization subject to a cardinality constraint. Their work is different from ours in that they adopted a different and more general metric of fairness, leading to a different optimization problem from ours. Due to its intrinsic hardness, there does not exist constant approximation algorithms for this problem. Moreover, their results only apply to the non-adaptive setting. Nevertheless, we also discuss their fairness notation under the adaptive setting within an extension of our research.  While the previously mentioned studies primarily concentrate on identifying a static set that adheres to \emph{rigid} fairness constraints, there has been a growing interest in identifying a random set that achieves fairness constraints on average \citep{tang2023beyond, yuan2023group, tang2023ctw}. Our work is also related to non-monotone submodular maximization. Similar to existing approaches \citep{tang2021pointwise, amanatidis2020fast}, we adopt sampling techniques to overcome challenges related to non-monotone objective functions.

We next discuss  related research on adaptive submodular optimization.  \cite{golovin2011adaptive} introduced the notation of \emph{adaptive submodularity}, extending the classic notation of submodularity from the non-adaptive setting to the adaptive setting. Their framework and its variants \citep{tang2021beyond,tang2021optimal,tang2021pointwise} can capture those scenarios whose objective function is stochastic and it can be learned as we select more items. Following this framework, we develop the first adaptive policy that achieves a constant approximation ratio against the best adaptive policy. To our knowledge, there is no existing studies for the adaptive setting that can handle the fairness constraints.

\section{Preliminaries and Problem Statement}
Throughout the paper, we use $[m]$ to denote the set $\{1, 2, \cdots, m\}$.
\subsection{Non-adaptive setting} Under the non-adaptive setting, the input of our problem is a set $V$ of $n$ items and a non-negative submodular utility function $f: 2^V\rightarrow \mathbb{R}_+$. Given two sets $X, Y\subseteq V$, we define the marginal utility of $X$ on top of $Y$ as
$f(X\mid Y) = f(X\cup Y)-f(Y)$. We say a function  $f: 2^V\rightarrow \mathbb{R}_+$ is submodular if for any two sets $X, Y\subseteq V$ such that $X\subseteq Y$ and any item $e \in V\setminus Y$, $f(e\mid Y) \leq f(e\mid X)$.

We assume that $V$ is partitioned into $m$ groups: $V_1, V_2,\cdots, V_m$. Let $\alpha\in[0, n]$ be a group equality constraint. The non-adaptive submodular maximization problem with group equality ($\textbf{P.0}$) is listed as follows.

 \begin{center}
\framebox[0.6\textwidth][c]{
\enspace
\begin{minipage}[t]{0.6\textwidth}
\small
$\textbf{P.0}$
$\max f(S)$
\textbf{subject to:}
$|S \cap V_i|-|S \cap V_j| \leq \alpha, \forall i, j\in[m]$.
\end{minipage}
}
\end{center}
\vspace{0.1in}

%Let $\mathcal{F}$ denote a set of feasible solutions under this fairness constraint, i.e.,
%\[\mathcal{F} = \{S\subseteq V : |S \cap V_a|-|S \cap V_b| \leq \alpha, \forall a, b\in[m]\}\]
In the above formulation, $\alpha$ captures the degree of group equality in a feasible solution. As discussed earlier, if we set $\alpha=0$, then any feasible solution must contain the same number of items from each group, hence, it achieves the highest degree of group equality; if we set $\alpha=n$, then there is no group equality constraints. We first provide the hardness result of $\textbf{P.0}$.

\begin{lemma}
\label{lem:wash}
Problem $\textbf{P.0}$ is NP-hard.
\end{lemma}

\subsection{Adaptive setting}
We next introduce our problem under a more complicated adaptive setting \citep{golovin2011adaptive}. On a high level, each item has a random state and the utility of an evaluated set is dependent on the state of all items from that set. However, the realized state of an item is unknown initially, one must select an item before observing its realization. Hence, a typical adaptive solution can be viewed as a sequence of selections and each selection depends on past outcomes. One example of this setting is experiment design, where the practitioner  performs a sequence of tests on a patient in order to reach the most accurate conclusion.

Formally, along with the set $V$, each item $e\in V$ has a random state $\Phi(e)$ drawn from $O$. We use $\phi(e)$ to denote a realization of $\Phi(e)$. Hence, a \emph{realization} $\phi$ can be considered as a mapping function that maps items to states: $\phi: V \rightarrow O$. In the example of experiment design,  an item $e$ represents a test, such as the blood pressure, and
$\Phi(e)\in\{\emph{high}, \emph{low}, \emph{normal}\}$ is the outcome of the test. We further assume that  the prior probability distribution $p(\phi) = \Pr(\Phi = \phi)$ over realizations $\phi$ is known. Given any set of items $S\subseteq V$, we use $\psi: S\rightarrow O$ to denote a \emph{partial realization} and $\mathrm{dom}(\psi)=S$ is the \emph{domain} of $\psi$. We say a realization $\phi$ is consistent with a partial realization $\psi$, denoted $\phi \succeq \psi$, if they are equal everywhere in the domain of $\psi$. We say a partial realization $\psi$  is a \emph{subrealization} of another partial realization $\psi'$, and denoted by $\psi \subseteq \psi'$, if $\mathrm{dom}(\psi) \subseteq \mathrm{dom}(\psi')$ and they are equal everywhere in  the domain of $\psi$. Given a partial realization $\psi$, let $p(\phi\mid \psi)$ denote the conditional distribution over realizations conditional on  $\psi$: $p(\phi\mid \psi) =\Pr[\Phi=\phi\mid \Phi\succeq \psi ]$.

A typical solution under the adaptive setting can be represented as  a policy $\pi$. Formally, a policy $\pi$ can be encoded as a mapping function that maps a set of partial realizations  to certain distribution $\mathcal{P}(V)$ of $V$: $\pi: 2^{V\times O} \rightarrow \mathcal{P}(V)$. It tells which item to select next based on the feedback (partial realization) from selected items. For instance, consider a scenario where we have chosen a set $S$ of items, and observe a partial realization $\cup_{e\in S}\{(e, \phi(e))\}$. If $\pi(\cup_{e\in S}\{(e, \phi(e))\})= e'$, then $\pi$ selects $e'$ as the next item.

There is a utility function $f : 2^{V\times O} \rightarrow \mathbb{R}_{\geq0}$ whose value is jointly decided by items and states.  For a policy $\pi$ and any realization $\phi$, let $V(\pi, \phi)$ denote the subset of items selected by $\pi$ under realization $\phi$. One can represent the expected  utility $f_{avg}(\pi)$ of a policy $\pi$ as
$
f_{avg}(\pi)=\mathbb{E}_{\Phi\thicksim p(\phi), \Pi}[f(V(\pi, \Phi), \Phi)]$,
 where the expectation is taken over $\Phi$ with respect to $p(\phi)$ and the internal randomness of $\pi$.

The conditional expected marginal utility $\Delta(e \mid \psi)$ of an item $e$ on top of a partial realization $\psi$ is defined as follows:
$
\Delta(e \mid  \psi)=\mathbb{E}_{\Phi}[f(\mathrm{dom}(\psi) \cup \{e\}, \Phi)-f(\mathrm{dom}(\psi), \Phi)\mid \Phi \succeq \psi]$,
where the expectation is taken over $\Phi$ with respect to $p(\phi\mid \psi)=\Pr(\Phi=\phi \mid \Phi \succeq \psi)$. The conditional expected marginal utility $\Delta(\pi \mid \psi)$ of a policy $\pi$ on top of a partial realization $\psi$ is defined as follows:
$
\Delta(\pi \mid  \psi)=\mathbb{E}_{\Phi}[f(V(\pi, \Phi) \cup \mathrm{dom}(\psi), \Phi)-f(\mathrm{dom}(\psi), \Phi)\mid \Phi \succeq \psi]$.

\begin{definition}[Adaptive Submodularity]
\label{def:122}A function $f : 2^{V\times O} \rightarrow \mathbb{R}_{\geq0}$ is adaptive submodular if for any two partial realizations $\psi$ and $\psi'$ such that $\psi\subseteq \psi'$, and for each $e\in V\setminus \mathrm{dom}(\psi')$,
$
\Delta(e\mid  \psi) \geq \Delta(e\mid  \psi')$.
\end{definition}

Now we are ready to introduce the adaptive submodular maximization problem with group equality. Given an adaptive submodular function $f : 2^{V\times O} \rightarrow \mathbb{R}_{\geq0}$ and a group equality constraint $\alpha$, our goal is to identify a best policy $\pi$ such that under every possible realization, $\pi$ selects a group of items that satisfies the group equality constraint. A formal definition of our problem can be found as follows:

 \begin{center}
\framebox[1\textwidth][c]{
\enspace
\begin{minipage}[t]{1\textwidth}
\small
$\textbf{P.1}$
$\max f_{avg}(\pi)$
\textbf{subject to:}  $|V(\pi, \phi) \cap V_i|-|V(\pi, \phi) \cap V_j| \leq \alpha, \forall i, j\in[m], \forall \phi \in\{\phi\in O^V: p(\phi)>0\}$.
\end{minipage}
}
\end{center}
\vspace{0.1in}

We will now present additional notations  that will find application in subsequent discussions. Given two policies $\pi$ and $\pi'$,  we use $\pi @\pi'$ to denote a policy that runs $\pi$ first, and then runs $\pi'$, ignoring the partial realization obtained from running $\pi$. For any partial realization $\psi$, let us define a function $g_{\psi} : 2^{V\setminus \mathrm{dom}(\psi)}\rightarrow \mathbb{R}_{\geq0} $ as follows: $g_{\psi}(S)=\mathbb{E}_{\Phi\succeq \psi}[f(\mathrm{dom}(\psi)\cup S, \Phi)]$ where the expectation is taken over $\Phi$ with respect to $p(\phi\mid \psi)=\Pr(\Phi=\phi \mid \Phi \succeq \psi)$. Intuitively, $g_{\psi}(S)$ represents the expected utility after adding $S$ to $\mathrm{dom}(\psi)$ conditional on $\psi$. We next show that $g_{\psi}$ is a submodular function. This property will be used later to analyze the performance of our proposed policy.

\begin{lemma}
\label{lem:g}
Suppose $f : 2^{V\times O} \rightarrow \mathbb{R}_{\geq0}$ is adaptive submodular. For any partial realization $\psi$, $g_{\psi} : 2^{V\setminus \mathrm{dom}(\psi)}\rightarrow \mathbb{R}_{\geq0} $  is a submodular function.
\end{lemma}

\subsection{Two Examples}
\label{sec:example}
We next present two applications of our problem.

\paragraph{Example 1:  Recommendation Systems.} Let us revisit the hierarchical recommender systems example discussed in the introduction section. In the context of Uber Eats, users are frequently presented with various groups of recommendations, such as ``Chinese Food'' or ``Thai Food.'' Each group of restaurants is showcased through carousels, enabling customers to horizontally browse through different dining options, and each carousel contains approximately the same number of restaurants, irrespective of the cuisine type. A potential objective for the platform could be to choose the most suitable collection of restaurants that satisfies group-equality constraints, while maximizing the anticipated conversion rate.  Formally, this problem takes a set $V$ of $n$ restaurants as input, which is divided into $m$  groups denoted by $V_1, V_2, \cdots, V_m$. In addition, there are $l$ customer types, and the proportion of customers belonging to type $j$ (where $j \in [l]$) is represented by $\theta_j$. Under the well-known mixed multinomial logit (Mixed-MNL) model, each product $i\in V$ has a preference weight $\nu_{ij}$ with respect to customer type $j\in[l]$ and let $\nu_{0j}$ denote the preference for no purchase with respect to customer type $j\in[l]$.  Given an assortment of recommendations $S\subseteq V$ and a customer of type $j\in[l]$, the conversion probability of any product $i$ (where $i \in S$) can be calculated as follows:
\begin{eqnarray}
\label{eq:r1}\frac{ \nu_{ij}}{\nu_{0j}+\sum_{i\in S} \nu_{ij}}. \end{eqnarray}

It follows that the expected conversion rate of offering $S$ is
\begin{eqnarray}
\label{eq:r2}
f(S)=\sum_{j\in[l]}\theta_j\cdot\frac{ \nu_{ij}}{\nu_{0j}+\sum_{i\in S} \nu_{ij}}.
 \end{eqnarray}

It is easy to verify that the function (\ref{eq:r1}) is a submodular function in terms of $S$, hence, the utility function (\ref{eq:r2}) is also submodular by the fact that a linear combination of submodular functions is still submodular. Our objective is to select an assortment of recommendations $S$ to maximize $f(S)$ such that $|S \cap V_i|-|S \cap V_j| \leq \alpha, \forall i, j\in[m]$. Since $f$ as defined in (\ref{eq:r2}) is a submodular function, this problem can be represented using the $\textbf{P.0}$ formulation.

\paragraph{Example 2: Seed Selection for Influence Maximization.} The idea of influence maximization is to select a group of influential individuals or seeds to help promote certain products or ideas through an online social platform. We capture the structure of the social network by a directed weighted graph. Each node represents a user and each edge represents the relationship between a pair of users. The goal is to select an initial set of nodes to maximize the spread of influence measured by the expected number of nodes that will ultimately be active according to the propagation model. We focus here on the independent cascade (IC) model, according to which, each edge in the graph is set to be \emph{live} independently with probability $p'$, otherwise it is \emph{blocked}. The influence can only propagate via \emph{live} edges.

For this problem, the ground set $V$ consists of network nodes. Assume $V$ is partitioned into $m$ groups: $V_1, V_2, \cdots, V_m$. Assume that each seed node incurs a unit cost, i.e. we have a cost term $f_{\text{cost}}(S) = |S|$, which results in the following objective: $f(S)=f_{\text{inf}}(S) - |S|$. Here $f_{\text{inf}}(S)$ is the expected number of nodes that can be reached from the seed set $S$ via \emph{live} edges. It is proved that $f_{\text{inf}}$ is monotone and submodular \citep{kempe2003maximizing}. Since the cost term is modular, $f(S)$ is still submodular. It is also non-negative, given $f_{\text{inf}}(S) \geq |S|$. Our objective is to find a group equality-aware set of seeds $S$ to maximize $f(S)$ such that $|S \cap V_i|-|S \cap V_j| \leq \alpha, \forall i, j\in[m]$.

In the adaptive version of the problem, we denote by $f_{\text{inf}}(S,\phi)$ the number of ultimately active nodes under realization $\phi$, if the nodes in $S$ are initially active. Here each realization $\phi$ corresponds to a full outcome of the IC model, that is, an assignment to each edge of being either \emph{live} or \emph{blocked}.  When a node $e\in V$ is selected, it reveals the status of all outgoing edges of $e$ and of any node that can be reached from $e$ via \emph{live} edges. Note for a policy $\pi$ and any realization $\phi$, we denote by $V(\pi,\phi)$ the subset of items selected by $\pi$ under realization $\phi$. Our objective becomes $f_{avg}(\pi)=\mathbb{E}_{\Phi,\Pi}[f(S,\Phi)]=\mathbb{E}_{\Phi,\Pi}[f_{\text{inf}}(S,\Phi)-|S|]$. Here $S=V(\pi, \phi)$ denotes the subset of nodes selected by $\pi$ under realization $\phi$. \cite{golovin2011adaptive} showed that $f_{\text{inf}}$ is monotone adaptive submodular, therefore $f_{avg}(\pi)$ is also adaptive submodular.

\section{Non-adaptive Submodular Maximization with Group Equality}
\label{sec:non-adaptive-general}
In this section, we study our problem under the non-adaptive setting. For simplicity, let $k_i=|V_i|$ denote the size of $V_i$ for each group $i\in[m]$. Let $k_{\min}=\min_{i\in[m]} k_i$ denote the size of the smallest group. Unless otherwise specified, we use $S_i$ to represent $S\cap V_i$ for any set $S\subseteq V$ and any $i\in[m]$.

\subsection{Algorithm Design}
\label{sec:algorithm-sg}
In this section, we introduce the design of our algorithm. Our algorithm is based on a simple  greedy algorithm which selects items based on their  marginal utility. However, because our utility function is non-monotone, simply selecting the item based on marginal utility could lead to traps of low utility. To this end, we add a sampling phase to our carefully designed greedy algorithm to avoid this trap and extend its
guarantees to the non-monotone case.  We next explain our algorithm in details. Our algorithm is composed of three phases:
\begin{enumerate}
\item We first select a random subset $R$ such that each item $e\in V$ is included in $R$ independently with probability $p\in[0,1]$. The value of $p$ will be optimized later.

\item Then we run a greedy algorithm \textsf{Greedy} only on $R$ to select a semi-feasible solution. Next we first introduce the concept of semi-feasibility, then explain \textsf{Greedy} in details.

 \begin{definition}
 \label{def:1}We call a set $S\subseteq V$ semi-feasible  if $|S \cap V_i|\leq \min\{\lfloor\frac{k_i}{2}\rfloor, \lfloor\frac{k_{\min}}{2}\rfloor+\alpha\}$ for all groups $i\in[m]$.
 \end{definition}

% On the high level, the greedy algorithm picks the item that has the largest marginal gain in each step subject to a simple partition matroid constraint.
\textsf{Greedy} starts with an empty set $A^{\textsf{greedy}}=\emptyset$. In each subsequent iteration, it finds an item with the largest marginal gain from $R$ such that adding that item to $A^{\textsf{greedy}}$ does not violate the semi-feasibility defined in Definition \ref{def:1}. If this marginal gain is positive, we add it to the current solution; otherwise, we terminate the algorithm and return  $A^{\textsf{greedy}}$. A detailed description of \textsf{Greedy} is listed in Algorithm \ref{alg:1}.
\item Note that $A^{\textsf{greedy}}$ is not necessarily a feasible solution to the original problem $\textbf{P.0}$. That is, there may exist some two groups $i,j\in[m]$ such that $|A^{\textsf{greedy}} \cap V_i|-|A^{\textsf{greedy}} \cap V_j| > \alpha$. We next explain how to obtain a feasible solution $A^{\textsf{final}}$ by adding some additional items to $A^{\textsf{greedy}}$. For each group $i\in[m]$ such that $|A^{\textsf{greedy}} \cap V_i|< \min\{\lfloor\frac{k_i}{2}\rfloor, \lfloor\frac{k_{\min}}{2}\rfloor+\alpha\}$, we first pick   two arbitrary sets $X_i$ and $Y_i$  from  $V_i\setminus A^{\textsf{greedy}}$ such that $|X_i|=|Y_i|=\min\{\lfloor\frac{k_i}{2}\rfloor, \lfloor\frac{k_{\min}}{2}\rfloor+\alpha\} - |A^{\textsf{greedy}} \cap V_i|$ and $X_i\cap Y_i =\emptyset$. We can always find such two disjoint sets due to the following observations: Because $|A^{\textsf{greedy}} \cap V_i|< \min\{\lfloor\frac{k_i}{2}\rfloor, \lfloor\frac{k_{\min}}{2}\rfloor+\alpha\}\leq \lfloor\frac{k_i}{2}\rfloor$, where the first inequality is due to our assumption, we have $|V_i\setminus A^{\textsf{greedy}}|\geq 2\cdot (\lfloor\frac{k_i}{2}\rfloor-|A^{\textsf{greedy}} \cap V_i|)$. This implies that  $|V_i\setminus A^{\textsf{greedy}}|\geq 2\cdot (\min\{\lfloor\frac{k_i}{2}\rfloor, \lfloor\frac{k_{\min}}{2}\rfloor+\alpha\} -|A^{\textsf{greedy}} \cap V_i|)$. Hence, $V_i\setminus A^{\textsf{greedy}}$ is large enough to contain two disjoint sets, each of which has size $\min\{\lfloor\frac{k_i}{2}\rfloor, \lfloor\frac{k_{\min}}{2}\rfloor+\alpha\} - |A^{\textsf{greedy}} \cap V_i|$.

    Let $L=\{i\in[m]\mid |A^{\textsf{greedy}} \cap V_i|< \min\{\lfloor\frac{k_i}{2}\rfloor, \lfloor\frac{k_{\min}}{2}\rfloor+\alpha\}\}$. Then we build two candidate final solutions $A^1$ and $A^2$ as follows:
    \[A^1=  A^{\textsf{greedy}}\cup\{\cup_{i\in L} X_i\}; A^2=  A^{\textsf{greedy}}\cup\{\cup_{i\in L} Y_i\}.\]
    Finally, we choose the better solution between $A^1$ and $A^2$ as the final solution $A^{\textsf{final}}$, that is, $f(A^{\textsf{final}})=\max\{f(A^1), f(A^2)\}$.
\end{enumerate}

\begin{algorithm}[hptb]
\caption{Greedy Algorithm}
\label{alg:1}
\begin{algorithmic}[1]
\STATE $R$ is a random set sampled from $V$, $A^{\textsf{greedy}}=\emptyset$
\WHILE {true}
%\STATE observe $\psi_i$;
\STATE let $e'=\argmax_{e\in R: e\cup A^{\textsf{greedy}} \mbox{ is semi-feasible }}f(e \mid A^{\textsf{greedy}})$
\IF {$f(e' \mid A^{\textsf{greedy}})>0$}
\STATE $A^{\textsf{greedy}} = A^{\textsf{greedy}}\cup\{e'\}$
\ELSE
\STATE break
\ENDIF
\ENDWHILE
\RETURN $A^{\textsf{greedy}}$
%\RETURN $S_k$
\end{algorithmic}
\end{algorithm}

\subsection{Performance Analysis}
\label{sec:performance}
 %Let $A^{\textsf{final}}$ denote the final solution returned from our algorithm, that is,  $A^{\textsf{final}}$ is the better solution between $A^1$ and $A^2$.
 We first prove that  $A^{\textsf{final}}$ is a feasible solution to our original problem.
 \begin{lemma}
 \label{lem:feasible}
  $A^{\textsf{final}}$ is a feasible solution to $\textbf{P.0}$.
 \end{lemma}

 To facilitate our analysis, we consider an alternative way of implementing our algorithm as follows. Instead of picking a random set $R$ at the beginning, we toss a coin of success $p$ to decide whether or not to select an item once this item is being considered. In other words, we integrate the sampling phase into the selection process. It is easy to verify that this change does not affect the output distribution of our algorithm.

Given the greedy solution $A^{\textsf{greedy}}$, let $W(A^{\textsf{greedy}})=\{e\in V\mid f(e\mid A^{\textsf{greedy}})>0\}$ denote the set of all items whose marginal utility with respect to $A^{\textsf{greedy}}$ is positive. For each $i\in[m]$, we number all items in $W(A^{\textsf{greedy}})\cap V_i$  by decreasing value of $f(\cdot \mid A^{\textsf{greedy}})$, i.e., $e^i_1\in \arg\max_{e\in W(A^{\textsf{greedy}})\cap V_i} f(e \mid A^{\textsf{greedy}})$. Let $l_i=\min\{|W(A^{\textsf{greedy}})\cap V_i|, k_i, k_{\min}+\alpha\}$. For each $i\in[m]$, define $D_i(A^{\textsf{greedy}})=\{e^i_q\in W(A^{\textsf{greedy}})\cap V_i \mid q\in[l_i]\}$ as the set containing the first $l_i$ items from $W(A^{\textsf{greedy}})\cap V_i$. Intuitively, $D_i(A^{\textsf{greedy}})$ contains a set of \emph{best-looking} items on top of $A^{\textsf{greedy}}$.
%If $l=1$, let $d(\psi)= \frac{B-c(e(1))}{c(e(1))} \Delta(e(1)\mid \mathrm{dom}(\psi), \psi)$; otherwise, if $l>1$, let
%\begin{eqnarray}
%d(\psi) = \sum_{j=1}^{l-1}  \Delta(e(j)\mid \mathrm{dom}(\psi), \psi) + \frac{B-\sum_{j=1}^{l-1} c(e(j))}{c(e(l))}\Delta(e(l)\mid \mathrm{dom}(\psi), \psi)
%\end{eqnarray}

%Given a $A^{\textsf{greedy}}$, for each $i\in[m]$, define \begin{eqnarray}
%d_i(A^{\textsf{greedy}}) = \sum_{e\in D_i(A^{\textsf{greedy}})} f(e \mid A^{\textsf{greedy}})~\nonumber
%\end{eqnarray}

Let $OPT$ denote the optimal solution of $\textbf{P.0}$. In analogy to Lemma 1 of \citep{gotovos2015non},
\begin{eqnarray}
\label{eq:ody}
\sum_{i\in[m]} \sum_{e\in D_i(A^{\textsf{greedy}})} f(e \mid A^{\textsf{greedy}}) \geq  f(OPT\mid A^{\textsf{greedy}}).
\end{eqnarray}

For each $i\in[m]$, define $C_i(A^{\textsf{greedy}})$ as those items in $D_i(A^{\textsf{greedy}})$ that have been considered by \textsf{Greedy} but not added to the solution because of the coin flips. Let $U_i(A^{\textsf{greedy}})$ denote those items in $D_i(A^{\textsf{greedy}})$ that have not been considered by \textsf{Greedy}. In the rest of this section, we drop the term $A^{\textsf{greedy}}$ from  $D_i(A^{\textsf{greedy}})$, $W(A^{\textsf{greedy}})$, $C_i(A^{\textsf{greedy}})$ and $U_i(A^{\textsf{greedy}})$ if it is clear from the context. (\ref{eq:ody}) can be rewritten as
\begin{eqnarray}
\label{eq:will}
\sum_{i\in[m]}\sum_{e\in C_i\cup U_i}f(e \mid A^{\textsf{greedy}}) \geq  f(OPT\mid A^{\textsf{greedy}}).
\end{eqnarray}

Now we are ready the analyze the approximation ratio of $A^{\textsf{final}}$. Note that because both $A^{\textsf{greedy}}$ and $A^{\textsf{final}}$ are some random sets, we focus on analyzing their expected performance. The outline of our analysis is as follows: We first analyze the expected performance bound of $A^{\textsf{greedy}}$ (Lemma \ref{lem:bash}, Lemma \ref{lem:summer}, Lemma \ref{lem:sharon-55}), and show that if we set $p=\frac{\sqrt{5}-1}{4}$, then $ \mathbb{E}_{A^{\textsf{greedy}}}[f(A^{\textsf{greedy}})] \geq 0.09\cdot f(OPT)$. Then in the proof of the main theorem (Theorem \ref{thm:1}),  we show that $\mathbb{E}_{A^{\textsf{final}}}[f(A^{\textsf{final}})]\geq \mathbb{E}_{A^{\textsf{greedy}}}[f(A^{\textsf{greedy}})]/2$. This, together with $\mathbb{E}_{A^{\textsf{greedy}}}[f(A^{\textsf{greedy}})] \geq 0.09\cdot f(OPT)$, implies that $\mathbb{E}_{A^{\textsf{final}}}[f(A^{\textsf{final}})]\geq 0.045\cdot f(OPT)$. In the rest of the analysis, we use $\mathbb{E}[f(A^{\textsf{greedy}})]$ and $\mathbb{E}[f(A^{\textsf{final}})]$   to denote $\mathbb{E}_{A^{\textsf{greedy}}}[f(A^{\textsf{greedy}})]$ and $\mathbb{E}_{A^{\textsf{final}}}[f(A^{\textsf{final}})]$ respectively.
\begin{lemma}
\label{lem:bash}
$
 \mathbb{E}[f(A^{\textsf{greedy}})]\geq p\cdot  \mathbb{E}[\sum_{i\in[m]}\sum_{e\in C_i}f(e \mid A^{\textsf{greedy}})]$.
\end{lemma}

\begin{lemma}
\label{lem:summer} Assume $k_{\min}> 1$, $ \mathbb{E}[f(A^{\textsf{greedy}})]\geq \frac{1}{3}\cdot  \mathbb{E}[\sum_{i\in[m]}\sum_{e\in U_i}f(e \mid A^{\textsf{greedy}})]$.
\end{lemma}

Now we are ready to analyze the expected performance bound of $A^{\textsf{greedy}}$.
\begin{lemma}
\label{lem:sharon-55}
If $k_{\min}> 1$ and we set $p=\frac{\sqrt{5}-1}{4}$, then
\begin{eqnarray}
\label{sharon-5}
\mathbb{E}[f(A^{\textsf{greedy}})] \geq 0.09\cdot f(OPT).
\end{eqnarray}

\end{lemma}

We are now in position to present the main theorem of this section.
\begin{theorem}
\label{thm:1}
If $k_{\min}> 1$ and we set $p=\frac{\sqrt{5}-1}{4}$, then
\begin{eqnarray}
\label{sharon-6}
\mathbb{E}[f(A^{\textsf{final}})] \geq 0.045 \cdot f(OPT).
\end{eqnarray}
\end{theorem}

\subsection{Solving the Case when $k_{\min} \leq 1$}
\label{sec:non-adaptive-special}
So far we assume that $k_{\min}> 1$,  now we are ready to tackle the case when $k_{\min}\leq 1$. In this case, $k_{\min}$ has two possible values: $0$ or $1$. We move this part to the online supplement (Section \ref{sec:non-adaptive-special-app}).

\subsection{Enhanced results for monotone case}
\label{sec:non-adaptive-monotone}
In this section, we show that if the utility function $f$ is monotone, then we can achieve a tight $(1-1/e)$-approximation ratio. Observe that if $f$ is monotone, then adding more items will never hurt the utility. Hence, we can simply select all items from each group whose size is $k_{\min}$. For the rest of the groups, we select at most $k_{\min}+\alpha$ items from each group.  Formally, we introduce the following optimization problem $\textbf{P.0.3}$. Let  $Z=\{i\in[m]\mid |V_i|=k_{\min}\}$ denote the set of the indexes of all smallest groups. The objective of  $\textbf{P.0.3}$ is $f'(\cdot)=f(\cdot\cup (\cup_{i\in Z} V_i))$, which is a monotone submodular function. Hence, $\textbf{P.0.3}$ is a classical monotone submodular maximization problem subject to a matroid constraint. There exists a tight $(1-1/e)$-approximation algorithm \citep{calinescu2007maximizing} for this problem. After solving this problem and obtain an output, we return this output together with $\cup_{i\in Z} V_i$ as the final solution.

 \begin{center}
\framebox[0.8\textwidth][c]{
\enspace
\begin{minipage}[t]{0.8\textwidth}
\small
$\textbf{P.0.3}$
$\max f'(S)$
\textbf{subject to:}
$S\subseteq V\setminus \cup_{i\in Z} V_i$; and
$ |S \cap V_i| \leq \alpha+k_{\min}, \forall i \in [m]\setminus Z$.
\end{minipage}
}
\end{center}
\vspace{0.1in}

\section{Adaptive Submodular Maximization with Group Equality}
\label{sec:adaptive-general}
In this section, we solve our problem under the more complicated adaptive setting. Our solution to  $\textbf{P.1}$ can be viewed as an adaptive variant of the algorithm proposed in the previous section. We next explain our policy $\pi^f$ in details.
\subsection{Design of $\pi^f$}
\label{sec:algorithm-asg}
\begin{enumerate}
\item We first select a random subset $R$ such that each item $e\in V$ is included in $R$ independently with probability $p$, where the value of $p$ will be optimized later.

\item Then we run an \emph{adaptive} greedy policy $\pi^g$ only on $R$. %At a high level, the adaptive greedy algorithm picks the item that has the largest marginal gain on top of the current realization $\psi$ subject to a simple partition matroid constraint listed in Definition \ref{def:1}.
 $\pi^g$ starts with an empty set $A^{\textsf{a-greedy}}=\emptyset$ and an empty observation $\psi_0=\emptyset$. In each subsequent iteration $t$, it finds an item $e'$ with the largest marginal gain on top of the current realization $\psi_{t-1}$ from $R$ such that adding $e'$ to $A^{\textsf{a-greedy}}$ does not violate the semi-feasibility defined in Definition \ref{def:1}, that is,
\begin{eqnarray}
e'=\argmax_{e\in R: e\cup A^{\textsf{a-greedy}} \mbox{ is semi-feasible }}\Delta(e \mid \psi_{t-1}).
\end{eqnarray}

If $\Delta(e_t \mid \psi_{t-1})$ is positive, then we add $e'$ to $A^{\textsf{a-greedy}}$ and update the partial realization using $\psi_t=\psi_{t-1}\cup\{e', \Phi(e')\}$; otherwise, we terminate the algorithm and return the current solution $A^{\textsf{a-greedy}}$. A detailed description of $\pi^g$ is listed in Algorithm \ref{alg:2}.
\item Note that $A^{\textsf{a-greedy}}$ does not necessarily satisfy the group equality constraint. That is, there may exist some two groups $i,j\in[m]$ such that $|A^{\textsf{a-greedy}} \cap V_i|-|A^{\textsf{a-greedy}} \cap V_j| > \alpha$. To create a feasible solution, we add some additional items to $A^{\textsf{a-greedy}}$ as follows. For each group $i\in[m]$ such that $|A^{\textsf{a-greedy}} \cap V_i|< \min\{\lfloor\frac{k_i}{2}\rfloor, \lfloor\frac{k_{\min}}{2}\rfloor+\alpha\}$, we first pick   two arbitrary sets $X_i$ and $Y_i$  from  $V_i\setminus A^{\textsf{a-greedy}}$ such that $|X_i|=|Y_i|=\min\{\lfloor\frac{k_i}{2}\rfloor, \lfloor\frac{k_{\min}}{2}\rfloor+\alpha\} - |A^{\textsf{a-greedy}} \cap V_i|$ and $X_i\cap Y_i =\emptyset$. We can always find such two sets because $|A^{\textsf{a-greedy}} \cap V_i|< \min\{\lfloor\frac{k_i}{2}\rfloor, \lfloor\frac{k_{\min}}{2}\rfloor+\alpha\}\leq \lfloor\frac{k_i}{2}\rfloor$, where the first inequality is due to our assumption. Let $L=\{i\in[m]\mid |A^{\textsf{a-greedy}} \cap V_i|< \min\{\lfloor\frac{k_i}{2}\rfloor, \lfloor\frac{k_{\min}}{2}\rfloor+\alpha\}$. We build two candidate final solutions $A^1$ and $A^2$ as follows:
    \[A^1=  A^{\textsf{a-greedy}}\cup(\cup_{i\in L} X_i); A^2=  A^{\textsf{a-greedy}}\cup(\cup_{i\in L} Y_i).\]

 Finally, $\pi^f$ chooses the better solution between $A^1$ and $A^2$ as the final solution $A^{\textsf{final}}$, that is, assuming $\psi$ is the partial realization of $A^{\textsf{a-greedy}}$, then $g_{\psi}(A^{\textsf{final}})=\max \{g_{\psi}(\cup_{i\in L} X_i), g_{\psi}(\cup_{i\in L} Y_i)\}$.
\end{enumerate}

\begin{algorithm}[hptb]
\caption{Adaptive Greedy Policy $\pi^g$}
\label{alg:2}
\begin{algorithmic}[1]
\STATE $R$ is a random set sampled from $V$, $A^{\textsf{a-greedy}}=\emptyset$, $t=0$, $\psi_0=\emptyset$
\WHILE {true}
%\STATE observe $\psi_i$;
\STATE let $e'=\argmax_{e\in R: e\cup A^{\textsf{a-greedy}} \mbox{ is semi-feasible }}\Delta(e \mid \psi_{t-1})$
\IF {$\Delta(e \mid \psi_{t-1})>0$}
\STATE $A^{\textsf{a-greedy}} = A^{\textsf{a-greedy}}\cup\{e'\}$; $\psi_t=\psi_{t-1}\cup\{(e', \Phi(e'))\}$; $t=t+1$
\ELSE
\STATE break
\ENDIF
\ENDWHILE
\RETURN $A^{\textsf{a-greedy}}$
%\RETURN $S_k$
\end{algorithmic}
\end{algorithm}

\subsection{Performance Analysis}
\label{sec:performance1}
 %Let $A^{\textsf{final}}$ denote the final solution returned from our algorithm, that is,  $A^{\textsf{final}}$ is the better solution between $A^1$ and $A^2$.
 We first show that  $\pi^f$ is a feasible policy to $\textbf{P.1}$.
 \begin{lemma}
$\pi^f$ is a feasible policy to $\textbf{P.1}$.
 \end{lemma}
To prove this lemma, it suffices to show that  $A^{\textsf{final}}$ satisfies the group equality constraint under every realization. We omit the proof because  for any given fixed realization, the same argument used to prove Lemma \ref{lem:feasible} can be used to prove the feasibility of $A^{\textsf{final}}$.

 For the purpose of analyzing the performance bound of   $\pi^f$, we consider an alternative way of sampling $R$ as described in Section \ref{sec:performance}. That is, we toss a coin of success $p$ to decide whether or not to select an item once this item is being considered. %Given the greedy solution $A^{\textsf{a-greedy}}$ and its realization $\psi^\lambda$, e.g., $\mathrm{dom}(\psi^\lambda)=A^{\textsf{a-greedy}}$,
 We define $\lambda = \{S^\lambda_{[e]}, \psi^\lambda_{[e]}\mid e\in S^\lambda\}$ as a fixed run of $\pi^g$, where  $S^\lambda$ contains all selected items under $\lambda$, $S^\lambda_{[e]}$ contains all items that are selected before $e$ is being considered, and $\psi^\lambda_{[e]}$ is the partial realization of $S^\lambda_{[e]}$. Hence, $S^\lambda$ is identical to $A^{\textsf{a-greedy}}$ for a fixed run $\lambda$. Moreover, we use $\psi^\lambda$ to denote the partial realization of  $S^\lambda$.  Let $W(\psi^\lambda)=\{e\in V\mid \Delta(e\mid \psi^\lambda)>0\}$ denote the set of all items whose marginal utility with respect to $\psi^\lambda$ is positive. For each $i\in[m]$, we number all items in $W(\psi^\lambda)\cap V_i=\{e^i_1, e^i_2, \cdots, e^i_{|W(\psi^\lambda)\cap V_i|} \}$  by decreasing value of $\Delta(\cdot \mid \psi^\lambda)$, i.e., $e^i_1\in \arg\max_{e\in W(\psi^\lambda)\cap V_i} \Delta(e\mid \psi^\lambda)$. Let $l_i(\psi^\lambda)=\min\{|W(\psi^\lambda)\cap V_i|, k_i, k_{\min}+\alpha\}$. For each $i\in[m]$, define $D_i(\psi^\lambda)=\{e^i_q\in W(\psi^\lambda)\cap V_i\mid q\in[l_i(\psi^\lambda)]\}$ as the set containing the first $l_i(\psi^\lambda)$ items from $W(\psi^\lambda)\cap V_i$. Intuitively, $D_i(\psi^\lambda)$ contains a set of \emph{best-looking} items on top of $\psi^\lambda$.
%If $l=1$, let $d(\psi)= \frac{B-c(e(1))}{c(e(1))} \Delta(e(1)\mid \mathrm{dom}(\psi), \psi)$; otherwise, if $l>1$, let
%\begin{eqnarray}
%d(\psi) = \sum_{j=1}^{l-1}  \Delta(e(j)\mid \mathrm{dom}(\psi), \psi) + \frac{B-\sum_{j=1}^{l-1} c(e(j))}{c(e(l))}\Delta(e(l)\mid \mathrm{dom}(\psi), \psi)
%\end{eqnarray}

%Fix a realization $\psi^\lambda$ of  $S^\lambda$, for each $i\in[m]$, define \begin{eqnarray}
%d_i(\psi^\lambda) = \sum_{e\in D_i(\psi^\lambda)} \Delta(e\mid \psi^\lambda)~\nonumber
%\end{eqnarray}

Let $\pi^*$ denote the optimal policy. Note that $\pi^*$ selects at most $k_{\min}+\alpha$ items from each group due to the group equality constraint. In analogy to Lemma 1 of \citep{gotovos2015non},
\begin{eqnarray}
\label{eq:ody-b}
\sum_{i\in[m]}\sum_{e\in D_i(\psi^\lambda)} \Delta(e\mid \psi^\lambda) \geq  \Delta(\pi^* \mid \psi^\lambda).
\end{eqnarray}

For each $i\in[m]$, let $C_i(\psi^\lambda)$ contain those items in $D_i(\psi^\lambda)$ that have been considered by $\pi^g$ but not added to the solution because of the coin flips. Let $U_i(\psi^\lambda)$ contain those items in $D_i(\psi^\lambda)$ that have not been considered by $\pi^g$. In the rest of this section, we drop the term $\psi^\lambda$ from $W(\psi^\lambda)$, $l_i(\psi^\lambda)$, $D_i(\psi^\lambda)$, $C_i(\psi^\lambda)$ and $U_i(\psi^\lambda)$ if it is clear from the context. (\ref{eq:ody-b}) can be rewritten as
\begin{eqnarray}
\label{eq:will-b}
\sum_{i\in[m]}\sum_{e\in C_i\cup U_i}\Delta(e\mid \psi^\lambda) \geq   \Delta(\pi^* \mid \psi^\lambda).
\end{eqnarray}

Now we are ready the analyze the approximation ratio of $\pi^f$. The outline of our analysis is as follows: We first analyze the performance bound of $\pi^g$ (Lemma \ref{lem:bash-b}, Lemma \ref{lem:summer-b}, Lemma \ref{lem:sharon-55-b}). In particular, we show that if we set $p=1/2$, then $f_{avg}(\pi^g) \geq f_{avg}(\pi^*)/12 $. Then in the proof of the main theorem (Theorem \ref{thm:1-b}),  we show that $f_{avg}(\pi^f) \geq f_{avg}(\pi^g)/2$. This, together with $f_{avg}(\pi^g) \geq  f_{avg}(\pi^*)/12$, implies that $f_{avg}(\pi^f)\geq  f_{avg}(\pi^*)/24$. Let $\mathcal{D}$ denote the distribution of $\lambda$ in the rest of the analysis.
\begin{lemma}
\label{lem:bash-b}
$f_{avg}(\pi^g) \geq p\cdot  \mathbb{E}_{\Lambda\sim \mathcal{D}}[\sum_{i\in[m]}\sum_{e\in C_i}\Delta(e \mid \psi^\Lambda)]$.
\end{lemma}

\begin{lemma}
\label{lem:summer-b}Assume $k_{\min}> 1$. $
f_{avg}(\pi^g) \geq \frac{1}{3}\cdot  \mathbb{E}_{\Lambda\sim \mathcal{D}}[\sum_{i\in[m]}\sum_{e\in U_i}\Delta(e \mid \psi^\Lambda)]$.
\end{lemma}

Based on the above two lemmas, we next provide a performance bound of $\pi^g$.
\begin{lemma}
\label{lem:sharon-55-b}
Assume $k_{\min}> 1$. If we set $p=1/2$, then
\begin{eqnarray}
\label{sharon-5-b}
f_{avg}(\pi^g) \geq \frac{1}{12} \cdot f_{avg}(\pi^*).
\end{eqnarray}

\end{lemma}

Now we are in position to provide the main theorem of this section.
\begin{theorem}
\label{thm:1-b}
Assume $k_{\min}> 1$. If we set $p=1/2$, then
\begin{eqnarray}
\label{sharon-6-b}
f_{avg}(\pi^f) \geq f_{avg}(\pi^*)/24.
\end{eqnarray}
\end{theorem}

\subsection{Solving the Case when $k_{\min}\leq 1$}
\label{sec:adaptive-special}
Now we are ready to examine the case when $k_{\min}\leq 1$. The case when $k_{\min}=0$ is trivial, because in this case we can select at most $\alpha$ items from each group. Hence, our problem is reduced to a standard adaptive submodular maximization problem subject to a partition matroid constraint formulated as follows:
 \begin{center}
\framebox[0.8\textwidth][c]{
\enspace
\begin{minipage}[t]{0.8\textwidth}
\small
$\textbf{P.1.1}$
$\max f_{avg}(\pi)$
\textbf{subject to:}
$\forall \phi$ with $p(\phi)>0$:
$|V(\pi, \phi) \cap V_i| \leq \alpha, \forall i \in[m]$.
\end{minipage}
}
\end{center}
\vspace{0.1in}

There exists a $1/6$-approximation algorithm \citep{tang2021pointwise} for this problem.

The rest of this section is devoted to addressing the case when $k_{\min}=1$. One possible approach to solving this problem is to generalize the solution proposed in Section \ref{sec:non-adaptive-special2} to the adaptive setting. This involves solving the problem for each guess of $\min_i|OPT_i|$ and selecting the best solution among these guesses. However, due to the restriction that previously selected items cannot be discarded in the adaptive framework, this approach is not feasible. As a result, we propose a ``guess-free'' solution that considers two subcases based on the value of $\alpha$.
\subsubsection{$k_{\min}=1$ and $\alpha=0$}
\label{sec:adaptive-special1}
If $k_{\min}=1$ and $\alpha=0$, then the optimal solution either selects nothing or selects exactly one item from each group. Because our utility function is non-negative, there must exist an optimal solution that selects exactly one item from each group. Let $T=\{i\in[m]\mid |V_i|=1\}$ denote the set of the indexes of those groups of size one. It is safe to add $\cup_{i\in T} V_i$ to our solution in advance, leading to a relaxed optimization problem listed in $\textbf{P.1.2}$.%, where
%\begin{eqnarray}
%f'_{avg}(\pi)=\mathbb{E}_{\Phi\sim p(\phi), \Pi}[f(V(\pi, \Phi)\cup (\cup_{i\in T} V_i), \Phi)].~\nonumber
%\end{eqnarray}

 \begin{center}
\framebox[0.9\textwidth][c]{
\enspace
\begin{minipage}[t]{0.9\textwidth}
\small
$\textbf{P.1.2}$
$\max f_{avg}(\pi)$ \\
\textbf{subject to:}
$\forall \phi$ with $p(\phi)>0$: $|V(\pi, \phi) \cap V_i| \leq 1, \forall i \in [m]\setminus T$ and
$|V(\pi, \phi) \cap V_i|  =1 , \forall i \in T$.
\end{minipage}
}
\end{center}
\vspace{0.1in}

Let $\pi^{P12}$ denote the optimal solution to $\textbf{P.1.2}$. It is easy to verify that
\begin{eqnarray}
\label{eq:eagles}
f(\pi^{P12})\geq f(\pi^*),
 \end{eqnarray}
 this is because $\pi^*$ is a feasible solution to $\textbf{P.1.2}$. However, $\pi^{P12}$ may not be a feasible solution of our original problem, e.g.,  $\pi^{P12}$ may select zero items from some groups and violate the group equality constraint. Next, we present a near-optimal policy $\pi'$ for $\textbf{P.1.2}$, then convert it a feasible policy of our original problem.

Before presenting the design of $\pi'$, we first introduce a new optimization problem $\textbf{P.1.3}(\phi(\cup_{i\in T} V_i))$ which takes an arbitrary partial realization $\phi(\cup_{i\in T} V_i)$ of $\cup_{i\in T} V_i$ as an input. The objective function of $\textbf{P.1.3}(\phi(\cup_{i\in T} V_i))$ is defined as
\begin{eqnarray}
f'_{avg}(\pi\mid \phi(\cup_{i\in T} V_i))=\mathbb{E}_{\Phi}[f(V(\pi, \Phi) \cup (\cup_{i\in T} V_i), \Phi)\mid \Phi \succeq \phi(\cup_{i\in T} V_i)].~\nonumber
\end{eqnarray}

 \begin{center}
\framebox[0.9\textwidth][c]{
\enspace
\begin{minipage}[t]{0.9\textwidth}
\small
$\textbf{P.1.3}(\phi(\cup_{i\in T} V_i))$
$\max f'_{avg}(\pi\mid \phi(\cup_{i\in T} V_i))$ \\
\textbf{subject to:}
$\forall \phi$ with $p(\phi)>0$:
$V(\pi, \phi)\subseteq \cup_{i\in [m]\setminus T} V_i$ and
$|V(\pi, \phi) \cap V_i| \leq 1, \forall i \in [m]\setminus T$.
\end{minipage}
}
\end{center}
\vspace{0.1in}

The goal of $\textbf{P.1.3}(\phi(\cup_{i\in T} V_i))$ is to find a policy that maximizes the expected utility on top of $\phi(\cup_{i\in T} V_i)$. It is easy to verify that if there exists a $d$-approximation policy for $\textbf{P.1.3}(\phi(\cup_{i\in T} V_i))$ for any $\phi(\cup_{i\in T} V_i)$, then there must exist a $d$-approximation policy for $\textbf{P.1.2}$.  Note that if $f(\cdot, \cdot): 2^{V\times O} \rightarrow \mathbb{R}_{\geq0}$ is adaptive submodular with respect to $p(\phi)$, then $f(\cdot \cup \mathrm{dom}(\psi), \cdot): 2^{V\times O} \rightarrow \mathbb{R}_{\geq0}$ must be adaptive submodular with respect to $p(\phi\mid \psi)$ for any $\psi$. Hence, $\textbf{P.1.3}(\phi(\cup_{i\in T} V_i))$  a classic adaptive submodular maximization problem subject to a partition matroid constraint. There exists a $1/6$-approximation algorithm \citep{tang2021pointwise} for this problem.

Now we are ready to present the design of $\pi'$.  $\pi'$ first selects all items from $\cup_{i\in T} V_i$ and observes their partial realization $\Phi(\cup_{i\in T} V_i)$; then it implements the $1/6$-approximation policy  \citep{tang2021pointwise} for $\textbf{P.1.3}(\Phi(\cup_{i\in T} V_i))$ to compute a solution $A$; finally, it returns $A^{P13}=A\cup (\cup_{i\in T} V_i)$ as the output.

Observe that $\pi'$ achieves an expected utility of at least $1/6$ fraction of the optimal solution, i.e.,
\begin{eqnarray}
\label{eq:shangxue}
f_{avg}(\pi') \geq \frac{1}{6} f_{avg}(\pi^{P12})\geq  \frac{1}{6} f_{avg}(\pi^*),
\end{eqnarray}
where the second inequality is due to (\ref{eq:eagles}). However, $\pi'$  is not necessarily a feasible policy of our original problem. For example, there may exist some $i\in [m]\setminus T$ such that $|A^{P13}\cap V_i|=0$, which violates the group equality constraint. We next convert $A^{P13}$ to a feasible solution by adding some additional items. Let $T'=\{i\in [m]\setminus T\mid |A^{P13}\cap V_i|=0\}$. By the definition of $T$, we have that for each $i\in [m]\setminus T$, we have $|V_i|\geq 2$. Hence, for each $i\in T'$, we have $|V_i|\geq 2$. We pick two arbitrary items, say $x_i$ and $y_i$, from each group $i\in T'$, and build two candidate solutions as follows:
\[A^1=  A^{P13}\cup(\cup_{i\in T'} \{x_i\}); A^2=  A^{P13} \cup(\cup_{i\in T'} \{y_i\}).\]
Finally, we choose the better solution between $A^1$ and $A^2$ as the final output . Following the same analysis conducted in the proof of Theorem \ref{thm:1-b}, we can show that the expected utility of this output is at least $f_{avg}(\pi') /2$. This, together with (\ref{eq:shangxue}), implies that our solution achieves an approximation ratio of $1/12$ for the original problem.

\subsubsection{$k_{\min}=1$ and $\alpha\geq 1$}
\label{sec:adaptive-special2}
 We first explain the design of our policy $\pi^{f1}$ for this case.
\begin{enumerate}
\item $\pi^{f1}$ first selects a random subset $R$ such that each item $e\in V$ is included in $R$ independently with probability $p$, where the value of $p$ will be optimized later.

\item Then it runs an \emph{adaptive} greedy policy only on $R$. %At a high level, the adaptive greedy algorithm picks the item that has the largest marginal gain on top of the current realization $\psi$ subject to a simple partition matroid constraint listed in Definition \ref{def:1}.
 Starts with an empty set $A=\emptyset$ and an empty observation $\psi_0=\emptyset$. In each subsequent iteration $t$, $\pi^{f1}$  finds an item $e'$ with the largest marginal gain on top of the current realization $\psi_{t-1}$ from $R$ such that adding $e'$ to $A$ does not violate the size constraint $\alpha$ of any group, that is,
\begin{eqnarray}
e'=\argmax_{e\in R: \forall i\in[m], |(\{e\}\cup A)\cap V_i|\leq \alpha}\Delta(e \mid \psi_{t-1}).
\end{eqnarray}

If $\Delta(e_t \mid \psi_{t-1})$ is positive, then add $e'$ to $A$ and update the partial realization using $\psi_t=\psi_{t-1}\cup\{e', \Phi(e')\}$; otherwise, return $A$ as the final solution, i.e., $A^{\textsf{final}}= A$.
\end{enumerate}

 The design of $\pi^{f1}$ is similar to that of $\pi^f$, however, $\pi^{f1}$ does not require an additional phase of converting $A$ to a feasible solution. This is because $A$  contains at most $\alpha$ items from each group by the design of $\pi^{f1}$, hence, $A$ must satisfy the group equality constraint.

 We next analyze the performance bound of $\pi^{f1}$. We first introduce some important notations (most of them are adapted from Section \ref{sec:performance1}).  We consider an alternative way of sampling $R$. That is, we toss a coin of success $p$ to decide whether or not to select an item once this item is being considered.  We define $\lambda = \{S^\lambda_{[e]}, \psi^\lambda_{[e]}\mid e\in S^\lambda\}$ as a fixed run of $\pi^{f1}$, where  $S^\lambda$ contains all selected items under $\lambda$, $S^\lambda_{[e]}$ contains all items that are selected before $e$ is being considered, and $\psi^\lambda_{[e]}$ is the partial realization of $S^\lambda_{[e]}$. Hence, $S^\lambda$ is identical to $A^{\textsf{final}}$ for a fixed run $\lambda$. Moreover, we use $\psi^\lambda$ to denote the partial realization of  $S^\lambda$.  Let $W(\psi^\lambda)=\{e\in V\mid \Delta(e\mid \psi^\lambda)>0\}$ denote the set of all items whose marginal utility with respect to $\psi^\lambda$ is positive. For each $i\in[m]$, we number all items in $W(\psi^\lambda)\cap V_i=\{e^i_1, e^i_2, \cdots, e^i_{|W(\psi^\lambda)\cap V_i|} \}$  by decreasing value of $\Delta(\cdot \mid \psi^\lambda)$, i.e., $e^i_1\in \arg\max_{e\in W(\psi^\lambda)\cap V_i} \Delta(e\mid \psi^\lambda)$. Let $l_i(\psi^\lambda)=\min\{|W(\psi^\lambda)\cap V_i|, k_i, \alpha+1\}$. For each $i\in[m]$, define $D_i(\psi^\lambda)=\{e^i_q\in W(\psi^\lambda)\cap V_i\mid q\in[l_i(\psi^\lambda)]\}$ as the set containing the first $l_i(\psi^\lambda)$ items from $W(\psi^\lambda)\cap V_i$. Intuitively, $D_i(\psi^\lambda)$ contains a set of \emph{best-looking} items on top of $\psi^\lambda$.

 For each $i\in[m]$, let $C_i(\psi^\lambda)$ contain those items in $D_i(\psi^\lambda)$ that have been considered by $\pi^g$ but not added to the solution because of the coin flips. Let $U_i(\psi^\lambda)$ contain those items in $D_i(\psi^\lambda)$ that have not been considered by $\pi^g$. We drop the term $\psi^\lambda$ from $W(\psi^\lambda)$, $l_i(\psi^\lambda)$, $D_i(\psi^\lambda)$, $C_i(\psi^\lambda)$ and $U_i(\psi^\lambda)$ if it is clear from the context.

Let $\pi^*$ denote the optimal policy. Note that $\pi^*$ selects at most $\alpha+1$ items from each group due to the group equality constraint and the assumption that $k_{\min}=1$. In analogy to (\ref{eq:ody-b}), we have
\begin{eqnarray}
\label{eq:will-b-1}
\sum_{i\in[m]}\sum_{e\in C_i\cup U_i}\Delta(e\mid \psi^\lambda) \geq   \Delta(\pi^* \mid \psi^\lambda).
\end{eqnarray}

Now we are in position to analyze the approximation ratio of $\pi^{f1}$. Following the same proof of Lemma \ref{lem:bash-b}, we have the following lemma.
\begin{lemma}
\label{lem:bash-b-1}
$
f_{avg}(\pi^{f1}) \geq p\cdot  \mathbb{E}_{\Lambda\sim \mathcal{D}}[\sum_{i\in[m]}\sum_{e\in C_i}\Delta(e \mid \psi^\Lambda)]$.
\end{lemma}

We next present the second technical lemma.

\begin{lemma}
\label{lem:summer-b-1}Assume $k_{\min}=1$ and $\alpha\geq 1$. $
f_{avg}(\pi^{f1}) \geq \frac{1}{2}\cdot  \mathbb{E}_{\Lambda\sim \mathcal{D}}[\sum_{i\in[m]}\sum_{e\in U_i}\Delta(e \mid \psi^\Lambda)]$.
\end{lemma}

Based on the above two lemmas, we next provide a performance bound of $\pi^{f1}$.
\begin{theorem}
\label{lem:sharon-55-b-1}
 Assume $k_{\min}=1$ and $\alpha\geq 1$. If we set $p=1/2$, then
\begin{eqnarray}
\label{sharon-5-b}
f_{avg}(\pi^{f1}) \geq \frac{1}{10} \cdot f_{avg}(\pi^*).
\end{eqnarray}

\end{theorem}

\subsection{Enhanced results for monotone case}
\label{sec:adaptive-monotone}
In this section, we show that if the utility function $f$ is adaptive monotone, that is $\forall e\in V, \psi: \Delta(e\mid  \psi)\geq 0$, then we can achieve a $1/2$-approximation ratio. Observe that if $f$ is monotone, then adding more items will never hurt the utility. Hence, we can simply select all items from each group whose size is $k_{\min}$. For the rest of the groups, we select at most $k_{\min}+\alpha$ items from each group adaptively. Formally, we introduce the following optimization problem $\textbf{P.1.4}$. Let  $Z=\{i\in[m]\mid |V_i|=k_{\min}\}$ denote the set of the indexes of all smallest groups. After selecting all items from $\cup_{i\in Z} V_i$, we observe their partial realization  $\Phi(\cup_{i\in Z} V_i)$. Then we solve the following $\textbf{P.1.4}(\Phi(\cup_{i\in Z} V_i))$, where
\begin{eqnarray}
f'_{avg}(\pi\mid \phi(\cup_{i\in Z} V_i))=\mathbb{E}_{\Phi}[f(V(\pi, \Phi) \cup (\cup_{i\in Z} V_i), \Phi)\mid \Phi \succeq \phi(\cup_{i\in Z} V_i)].~\nonumber
\end{eqnarray}

 \begin{center}
\framebox[0.9\textwidth][c]{
\enspace
\begin{minipage}[t]{0.9\textwidth}
\small
$\textbf{P.1.4}(\phi(\cup_{i\in Z} V_i))$
$\max f'_{avg}(\pi\mid \phi(\cup_{i\in Z} V_i))$ \\
\textbf{subject to:}
$\forall \phi$ with $p(\phi)>0$: $V(\pi, \phi)\subseteq \cup_{i\in [m]\setminus Z} V_i$ and
$|V(\pi, \phi) \cap V_i| \leq k_{\min}+\alpha, \forall i \in [m]\setminus Z$.
\end{minipage}
}
\end{center}
\vspace{0.1in}  Because $f$ is adaptive monotone, $\textbf{P.1.4}$ is a classical monotone adaptive submodular maximization problem subject to a matroid constraint. There exists a $1/2$-approximation algorithm \citep{golovin2011adaptive1} for this problem. After solving this problem and obtain an output, we return this output together with $\cup_{i\in Z} V_i$ as the final solution.

\section{Empirical Evaluation}

In this section, we empirically assess our proposed algorithms in the context of influence maximization. A detailed description of this example can be found in Section \ref{sec:example}. Our evaluation focuses on measuring the performance through expected utility of solutions, under both non-adaptive and adaptive scenarios using real-world large-scale datasets. We investigate diverse parameter settings and item grouping methods, examining their influence on solution quality. Additionally, we explore the impact of varying the value of $\alpha$, affirming its role as a threshold variable capturing group equality in feasible solutions. Our algorithms are implemented in Java and experiments are conducted on a Linux server with an Intel Xeon 2.40GHz CPU and 128GB memory.  The source codes of this work are available for public use \footnote{\url{https://github.com/j-yuan/GEquality}}.
\subsection{Experimental Setup}

\begin{table}
\centering
  \caption{Statistics of $k_{\min}$ under different grouping strategies}
  \label{tab:kmin}
    \begin{adjustbox}{width=1\textwidth}
\small
  \begin{tabular}{c|ccccc|ccccc}
    \toprule
    \texttt{Grouping strategy} & \multicolumn{5}{c|}{\texttt{Random}} & \multicolumn{5}{c}{\texttt{Gaussian}}\\
    \midrule
   	\texttt{Number of groups} & $3$ & $5$ & $7$ & $9$ & $10$ & $3$ & $5$ & $7$ & $9$ & $10$\\
   	\midrule
     \texttt{Average ${k}_{\min}$} & 2672 & 1601 & 1142 & 887 & 798 & 1139 & 770 & 578 & 452 & 409\\
    	 \midrule
    	 \texttt{Range of $k_{\min}$} & [2517, 2821] & [1433, 1764] & [1036, 1245] & [793, 981] & [712, 889] & [907, 1187] & [575, 806] & [416, 601] & [321, 472] & [278, 425]\\
  \bottomrule
\end{tabular}
\end{adjustbox}
\end{table}

\textbf{Datasets.} We run our experiments on \emph{Wikivote}, a large-scale benchmark social network widely used in the social computing literature.  \emph{Wikivote} contains $103,663$ votes from $8,066$ users participating in the elections from the Wikipedia community. Each node represents a user and an edge exists between a pair of nodes if one user votes for the other.

\textbf{Grouping Strategies.} We consider two grouping strategies to partition the nodes into separate groups. First we consider a random group assignment strategy. Given the number of groups $m$, an integer is sampled randomly from $[0, m)$ for each node as its group id. Then we consider a Gaussian-based group assignment strategy that captures the group membership imbalance in practice. Suppose we have $m$ groups of nodes in the ground set $V$. We obtain $X$, a set of $|V|$ numbers drawn from a Gaussian distribution with $\mu=\frac{m}{2}$. For each number $x\in X$, we assign $\left\lfloor x \right\rfloor$ as the group id for a node in $V$. The statistics of the value of $k_{\min}$ under different grouping strategies are summarized in Table \ref{tab:kmin}.

\textbf{Algorithms.} We evaluate the performance of our non-adaptive sampling greedy algorithm (SG) and adaptive sampling greedy algorithm (ASG), as described in Section \ref{sec:algorithm-sg} and \ref{sec:algorithm-asg} respectively. We also implement two heuristic algorithms as our benchmarks for comparison purpose. Heuristic with Interval algorithm (HI) is our non-adaptive benchmark.  HI first employs a sampling-based non-adaptive greedy algorithm (as described in Section \ref{sec:algorithm-sg}) to obtain a semi-feasible solution $A^{HI}$ such that for each group $i\in[m]$, it holds that $|A^{HI}\cap V_i|<k_{\min}+\alpha$. Then HI adds some additional nodes to $A^{HI}$ to ensure that the number of nodes selected from each group is within the interval of $[k_{\min}, k_{\min}+\alpha]$. In addition to HI, we have implemented the Adaptive Heuristic with Interval algorithm (AHI) as an adaptive benchmark. AHI shares similarities with HI, but it incorporates a sampling-based adaptive greedy algorithm (as described in Section \ref{sec:algorithm-asg}) instead of the non-adaptive greedy algorithm to find a semi-feasible solution.

\textbf{Parameter Settings.} In our experiments, we study the impact of varying group equality threshold $\alpha$ and that of varying number of groups $m$ ranging from $2$ to $20$. We adopt the IC model as diffusion model and assign a probability of $p'=0.01$ to each edge. We also vary the value of $p'$ and explore its impact on the quality of the solution. For our proposed algorithms SG and ASG, we set their independent sampling rate $p=0.9$. We measure the utility of SG through Monte Carlo simulation. For ASG, we measure the conditional marginal utility as the expected increase in utility based on the observations of the actual influence spread triggered by the current seed set. For each set of experiments, we evaluate the expected utility of the output with $1,000$ rounds of simulation and report the average results in the following subsection.

\begin{figure*}[hptb]
\centering
%\hspace*{-0.1in}
\includegraphics[scale=0.45]{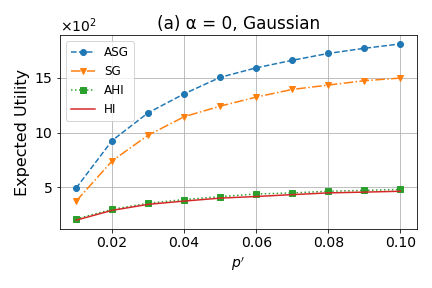}
\includegraphics[scale=0.45]{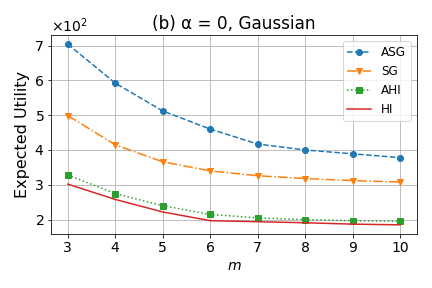}
\includegraphics[scale=0.45]{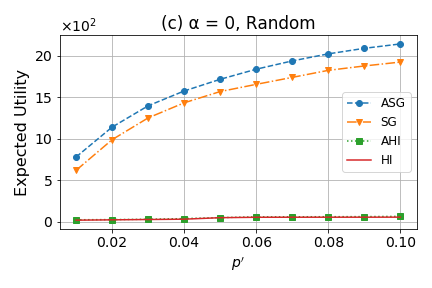}
\includegraphics[scale=0.45]{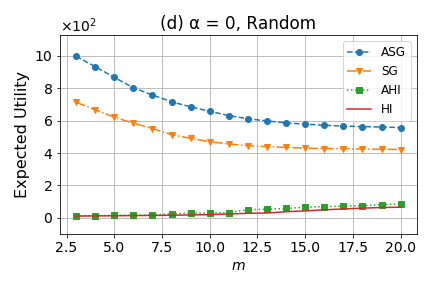}
\caption{The expected utility achieved by algorithms under different grouping strategies for $\alpha=0$}
\label{fig:alpha0}
\end{figure*}

\begin{figure*}[hptb]
\centering
%\hspace*{-0.1in}
\includegraphics[scale=0.45]{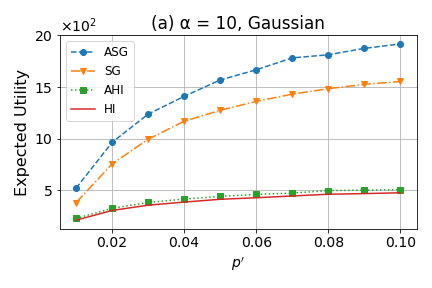}
\includegraphics[scale=0.45]{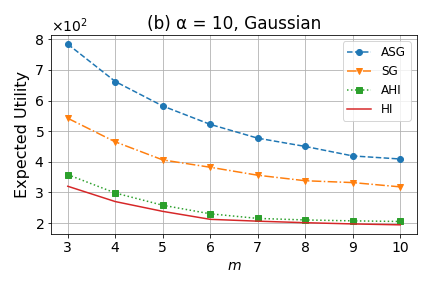}\\
\includegraphics[scale=0.45]{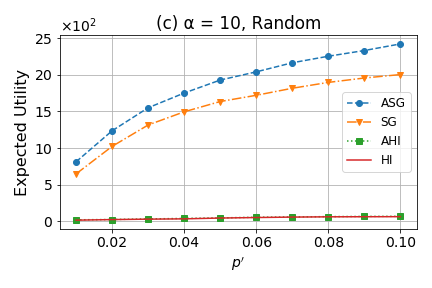}
\includegraphics[scale=0.45]{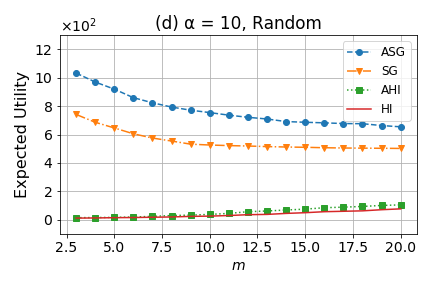}
\caption{The expected utility achieved by algorithms under different grouping strategies for $\alpha=10$}
\label{fig:alpha10}
\end{figure*}

\begin{figure*}[hptb]
\centering
%\hspace*{-0.39in}
\includegraphics[scale=0.45]{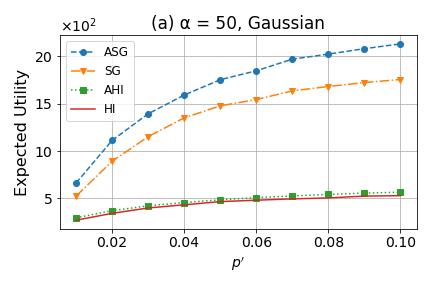}
\includegraphics[scale=0.45]{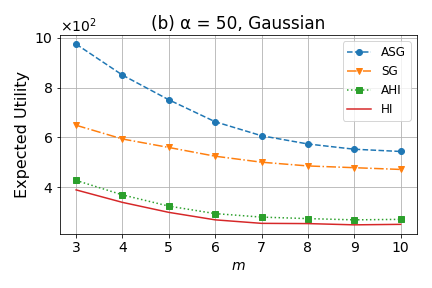}\\
\includegraphics[scale=0.45]{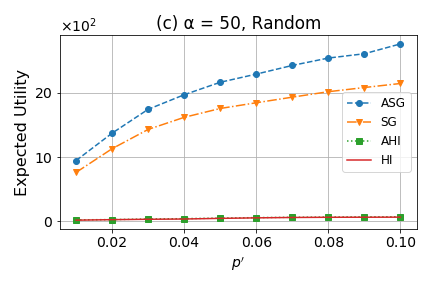}
\includegraphics[scale=0.45]{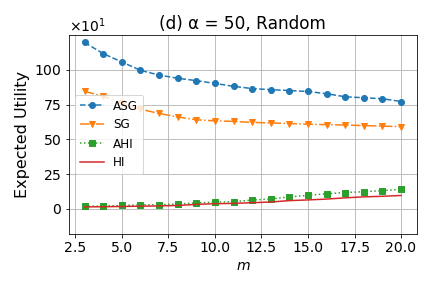}
\caption{The expected utility achieved by algorithms under different grouping strategies for $\alpha=50$}
\label{fig:alpha50}
\end{figure*}

\subsection{Experimental Results}

We compare the performance of ASG and SG with the benchmarks and report the results in Figure \ref{fig:alpha0}, \ref{fig:alpha10} and \ref{fig:alpha50} for $\alpha=0$, $\alpha=10$ and $\alpha=50$, respectively. $\alpha$ captures the degree of group equality in a feasible solution, and a smaller $\alpha$ indicates a higher degree of group equality. We also evaluate the performance of the algorithms under various settings of edge weight and number of groups with different grouping strategies.

We present in Figure \ref{fig:alpha0}(a) and \ref{fig:alpha0}(b) the results obtained under the Gaussian-based grouping strategy for $\alpha=0$. Figure \ref{fig:alpha10}(a) plots the expected utility yielded by the considered algorithms with respect to changes in the value of edge weight $p'$, ranging from $0.01$ to $0.1$. Here we set the number of groups $m=6$. As expected, as the edge weight increases, the expected utility also increases, since a higher edge weight enables more \emph{live} edges, leading to more nodes that are ultimately activated by the seed set. We observe that ASG and SG significantly outperform their benchmarks. ASG performs the best among all algorithms, and it outperforms SG by at least $20\%$ in terms of expected utility. This result verifies the superiority of our proposed algorithms.

Figure \ref{fig:alpha0}(b) illustrates the expected utility produced by the algorithms with respect to changes in the number of groups $m$, ranging from $3$ to $10$. Here we set the edge weight $p'=0.01$. We observe that the expected utility decreases as $m$ increases. The underlying reason is that under gaussian-based group strategy, a larger number of groups indicates that the smallest group has a smaller size. This leads to a smaller semi-feasible seed set found in the intermediate step of our algorithms, resulting in a smaller number of nodes that are ultimately activated by the seed set. Again, ASG outperforms SG by more than $20\%$ in terms of expected utility, and both of them significantly outperform their benchmarks.

We report in Figure \ref{fig:alpha0}(c) and \ref{fig:alpha0}(d) the results obtained under the random grouping strategy. The expected utility achieved by the proposed algorithms increases with $p'$, and decreases as $m$ increases. As shown in Figure \ref{fig:alpha0}(d), the expected utility of the benchmarks slightly increases with $m$ under the random grouping strategy. This is because when $m$ goes up, the average $k_{\min}$ goes down. In the test range, the benchmarks add less additional nodes on average to meet the lower bound ($k_{\min}$) for the number of nodes selected from each group, leading to a better expected utility. In addition, ASG performs the best among all three algorithms, and it outperforms SG by at least $10\%$ in terms of expected utility. Both ASG and SG outperform the benchmarks, this again demonstrates the efficacy of our proposed algorithms.

Moreover, we observe that compare with the random grouping strategy, the proposed algorithms produce a lower expected utility under the Gaussian-based grouping strategy, when all other settings are the same. The underlying reason is that for a fixed number of groups, with Gaussian-based grouping strategy, we end up with groups of disparate sizes. Gaussian-based grouping strategy is able to capture the imbalance in the nature of groups in practice. Our equality constraint based algorithms can ensure smaller groups receive a fair allocation of the resources (a.k.a. seeds).

Finally, we examine the impact of different values of $\alpha$ on the quality of the solution.  Figure \ref{fig:alpha10} and \ref{fig:alpha50} present the results for $\alpha=10$ and $\alpha=50$, respectively, and all other settings are the same as shown in Figure \ref{fig:alpha0}. Note $\alpha$ captures the degree of group equality in a feasible solution. A smaller $\alpha$ indicates a higher degree of group equality. We observe that the algorithms produce a lower expected utility when $\alpha$ is smaller. We consider this as the price of fairness as with a smaller $\alpha$, we ensure more equally allocated resources among groups at the cost of a larger degradation in the expected utility. In addition, we observe that ASG outperforms SG by more than $15\%$ in expected utility, both outperforming their benchmarks across various settings. This demonstrates the power of our adaptive strategy for the problem of submodular maximization with group equality constraints.

\section{Extension of Non-adaptive Case: Incorporating Global Cardinality Constraint}
\label{sec:non-adaptive-extended}

In this section, we consider a extended version of $\textbf{P.0}$ by incorporating a global cardinality constraint. A formal definition of this problem is listed in $\textbf{P.2}$. Our objective is to find a best $S$ such that it satisfies the group equality constraint $\alpha$ and a cardinality constraint $c$.
 \begin{center}
\framebox[0.8\textwidth][c]{
\enspace
\begin{minipage}[t]{0.8\textwidth}
\small
$\textbf{P.2}$
$\max f(S)$
\textbf{subject to:}
$|S \cap V_i|-|S \cap V_j| \leq \alpha, \forall i, j\in[m]$ and
$|S|\leq c$.
\end{minipage}
}
\end{center}
\vspace{0.1in}

Before presenting our solution, we first provide some useful observations. These observations will be used later to design and analyze our algorithm.

\subsection{Preliminaries}
\label{sec:pre}
In what follows, we show that there exists a solution $M$ such that
\begin{enumerate}
\item $f(M)\geq \frac{\kappa}{1+\kappa} f(OPT)$, where $\kappa=\min_{i\in [m]}\frac{|M\cap V_i|}{|OPT\cap V_i|}$.
\item For each $i\in[m]$, $|M\cap V_i|=\lfloor\frac{|OPT\cap V_i|}{2}\rfloor$.
\item Let $z=\min_{i\in[m]}|M\cap V_i|$,  each $i\in[m]$ satisfies $z \leq |M\cap V_i|\leq z+\alpha$, and $|M|\leq c$.
\end{enumerate}

The proof of the existence of such $M$ is deferred to the online supplement (Section \ref{sec:proof-of-m-app}).

\subsection{Algorithm Design}

Assuming the existence of the aforementioned $M$, and considering that we have the value of $z=\min_{i\in[m]}|M\cap V_i|$ (note that this assumption will be eliminated later), we proceed to introduce a new optimization problem denoted as $\textbf{P.2.1}$:

 \begin{center}
\framebox[0.8\textwidth][c]{
\enspace
\begin{minipage}[t]{0.8\textwidth}
\small
$\textbf{P.2.1}$
$\max f(S)$
\textbf{subject to:}
for each $i\in[m]$, $z \leq |S\cap V_i|\leq z+\alpha$ and
 $|S|\leq c$.
\end{minipage}
}
\end{center}
\vspace{0.1in}

The following lemma builds a quantitative relationship between the optimal solution to $\textbf{P.2.1}$ and the optimal solution to our original problem $\textbf{P.2}$.
\begin{lemma}
\label{lem:hospital1}
Every feasible solution to $\textbf{P.2.1}$ must be feasible to $\textbf{P.2}$. Let $S^{P21}$ be the optimal solution to $\textbf{P.2.1}$, we have that  $f(S^{P21})\geq \frac{\kappa}{1+\kappa} f(OPT)$, where $\kappa=\min_{i\in [m]}\frac{|M\cap V_i|}{|OPT\cap V_i|}$.
\end{lemma}

Lemma \ref{lem:hospital1} implies that if we can obtain an approximate solution to  $\textbf{P.2.1}$, then this solution is also an approximate solution (with a loss of $\frac{\kappa}{1+\kappa}$ factor in the approximation ratio) to our original problem $\textbf{P.2}$. Hence, in the rest of this section, we focus on solving $\textbf{P.2.1}$. Towards this end, we introduce another optimization problem $\textbf{P.2.2}$ as follows:
 \begin{center}
\framebox[0.9\textwidth][c]{
\enspace
\begin{minipage}[t]{0.9\textwidth}
\small
$\textbf{P.2.2}$
$\max f(S)$
\textbf{subject to:}
for each $i\in[m]$, $|S\cap V_i|\leq z+\alpha$ and
$\sum_{i\in[m]}\max\{z, |S\cap V_i|\} \leq c$.
\end{minipage}
}
\end{center}
\vspace{0.1in}

It is easy to verify that $\textbf{P.2.2}$ is a relaxation of $\textbf{P.2.1}$, that is, every feasible solution to $\textbf{P.2.1}$ must also be feasible to  $\textbf{P.2.2}$. Hence, the following lemma holds.

\begin{lemma}
\label{lem:hospital}
Let $S^{P22}$ be the optimal solution to $\textbf{P.2.2}$, we have $f(S^{P22})\geq f(S^{P21})$.
\end{lemma}

Note that not every feasible solution to $\textbf{P.2.2}$ is feasible to $\textbf{P.2.1}$. In particular, a feasible solution to $\textbf{P.2.2}$ does not necessarily satisfy the lower bound constraint in  $\textbf{P.2.1}$. Fortunately, we can make it feasible at the cost of losing a constant-factor in the approximation ratio by adding some additional items to it. It is worth noting that the constraint listed in $\textbf{P.2.2}$ satisfies the properties of a matroid constraint \citep{el2020fairness}. As a result, $\textbf{P.2.2}$ can be interpreted as a maximization problem involving a submodular function subject to a matroid constraint. Notably, there exists a $\frac{1}{e}-o(1)$-approximation randomized algorithm \citep{feldman2011unified} and a $0.283-o(1)$-approximation deterministic algorithm \citep{sun2023improved} that can be applied to this problem.

Now we are ready to present the design of our algorithm. The basic idea of our algorithm is to first find a solution to $\textbf{P.2.2}$, then convert it to a feasible solution to $\textbf{P.2.1}$. Our algorithm is composed of two phases:
\begin{enumerate}
\item We first call the randomized  algorithm in \citep{feldman2011unified} to solve $\textbf{P.2.2}$ to obtain a $\frac{1}{e}-o(1)$-approximation solution $A^{P22}$.

\item As discussed earlier, $A^{P22}$ may not satisfy the lower bound  of  $\textbf{P.2.1}$. To make it feasible, we add some additional items to $A^{P22}$ as follows. For each group $i\in[m]$ such that $|A^{P22} \cap V_i|< z$, we pick   two arbitrary sets $X_i$ and $Y_i$  from  $V_i\setminus A^{P22}$ such that $|X_i|=|Y_i|=z-|A^{P22} \cap V_i|$ and $X_i\cap Y_i =\emptyset$. Note that we can always find such two sets because $|A^{P22} \cap V_i|< z\leq \lfloor\frac{k_i}{2}\rfloor$, where the first inequality is due to our assumption and the second inequality is due to  $z=\min_{j\in[m]}|M\cap V_j|=\min_{j\in [m]}\lfloor\frac{|OPT_j|}{2}\rfloor\leq \min_{j\in[m]}\lfloor\frac{k_j}{2}\rfloor$.  Let $L=\{i\in[m]\mid |A^{P22} \cap V_i|< z\}$. Then we build two candidate solutions $A^1$ and $A^2$ as follows:
    \[A^1=  A^{P22}\cup(\cup_{i\in L} X_i); A^2=  A^{P22}\cup(\cup_{i\in L} Y_i).\]
    Finally, we choose the better solution between $A^1$ and $A^2$ as the final solution $A^{\textsf{final}}$, i.e., $f(A^{\textsf{final}})=\max\{f(A^1), f(A^2)\}$.
\end{enumerate}

\paragraph{\textbf{Remark 1:}} So far we assume that we know the value of $z$, to complete our algorithm design, we next discuss how to find out $z$ effectively. Because $z=\min_{j\in[m]}|M\cap V_j|= \min_{j\in[m]}\lfloor\frac{|OPT_j|}{2}\rfloor$, we have $z\leq  \min_{j\in[m]}\lfloor\frac{k_i}{2}\rfloor$. To find out $z$, we can simply enumerate all possibilities in the range of $[1, \min_{j\in[m]}\lfloor\frac{k_i}{2}\rfloor]$, and return the best one as the final solution.

\subsection{Performance Analysis}
 We first prove the feasibility of  $A^{\textsf{final}}$.
 \begin{lemma}
 \label{lem:jiaoshi}
 $A^{\textsf{final}}$ is a feasible solution to $\textbf{P.2.1}$ and hence $\textbf{P.2}$.
 \end{lemma}

We next analyze the approximation ratio of $A^{\textsf{final}}$. In the following theorem, we show that if  $\min_{i\in [m]}|OPT_i|> 1$, then $A^{\textsf{final}}$ achieves an approximation ratio of $\frac{1/e-o(1)}{8}$ in expectation, where the randomness is from  $A^{P22}$.
In Section \ref{sec:extended-speical}, we demonstrate that the assumption $\min_{i\in [m]}|OPT_i|> 1$ can be removed without impacting the approximation ratio.
 \begin{theorem}
 \label{lem:yao}
Let $OPT$ denote the optimal solution to $\textbf{P.2}$, assume $\min_{i\in [m]}|OPT_i|> 1$, $\mathbb{E}[f(A^{\textsf{final}})]\geq \frac{1/e-o(1)}{8}\cdot f(OPT)$.
 \end{theorem}

\paragraph{\textbf{Remark 2:}} We note that as compared with the performance bound derived in Theorem \ref{thm:1}, we achieve a better approximation ratio under the extended model (Theorem \ref{lem:yao}). However, one limitation of our proposed solution for the extended model is that it is not compatible with the adaptive framework. As noted in Remark 1, in order to implement this solution, we must try all possible values of $z$ and return the best solution among all guesses. Unfortunately, we can not afford such ``enumeration''  in the adaptive framework given that we are not allowed to discard any previously selected items in this setting. In this sense, our proposed solution under the basic model provides better robustness as it can be easily modified to achieve a good approximation guarantee in the adaptive setting.

\paragraph{\textbf{Remark 3:}} The approximation ratio presented in Theorem \ref{lem:yao} is in expectation. It is possible to de-randomize our algorithm by utilizing the deterministic algorithm proposed in \citep{sun2023improved} during phase 1 to solve $\textbf{P.2.2}$ and achieve a worst-case approximation of $0.283-o(1)$. By following the same proof outlined in Theorem \ref{lem:yao}, we can show that this algorithm attains a worst-case approximation of $\frac{0.283-o(1)}{8}$ for the original problem.

\subsection{Solving the case when $\min_{i\in [m]}|OPT_i|\leq 1$}
 \label{sec:extended-speical}
 Now we are ready to discuss the case when $\min_{i\in [m]}|OPT_i|> 1$ does not hold.  Observe that if this condition does not hold, then $\min_{i\in [m]}|OPT_i|= 0$ or $1$. We develop a  $\frac{1}{e}-o(1)$-approximation algorithm and a $\frac{1/e-o(1)}{2}$-approximation algorithm for these two cases, respectively. Although we do not know $\min_{i\in[m]}|OPT_i|$ initially, we can guess its value and solve the problem for each guess. Finally, the best solution among all guesses, including the one derived in the previous section, is returned as the final output. This, together with Theorem \ref{lem:yao}, indicates that this solution achieves an approximation ratio of $\min\{\frac{1/e-o(1)}{8}, \frac{1}{e}-o(1) , \frac{1/e-o(1)}{2}\}=\frac{1/e-o(1)}{8}$. We move this part to the online supplement (Section \ref{sec:extended-speical-app}).

\subsection{Enhanced results for monotone case}
\label{sec:extended-monotone}
We next show that if the utility function $f$ is monotone, then we can achieve a $1/2$-approximation ratio. We move this part to the online supplement (Section \ref{sec:extended-monotone-app}).

\section{Discussion on  Equity-fairness Constraints}
\label{sec:extended-equity}
Next, we discuss another frequently employed fairness notation referred to as ``equity-fairness constraints.'' Our focus here is to dynamically select a set of at most $c$ items in order to optimize an adaptive submodular function. This optimization is carried out while ensuring that the quantity of selected items from each group $V_i$ falls within the specified range of $[\underline{\gamma_i},\overline{\gamma_i}]$. A formal description of this problem is listed in $\textbf{P.3}$.
 \begin{center}
\framebox[0.97\textwidth][c]{
\enspace
\begin{minipage}[t]{0.97\textwidth}
\small
$\textbf{P.3}$
$\max f_{avg}(\pi)$
\textbf{subject to:}  \\
$\underline{\gamma_i} \leq |V(\pi, \phi) \cap V_i| \leq \overline{\gamma_i}, \forall i \in [m], \forall \phi \in\{\phi\in O^V: p(\phi)>0\}$ and
$ |V(\pi, \phi)| \leq c, \forall \phi \in\{\phi\in O^V: p(\phi)>0\}$.
\end{minipage}
}
\end{center}
\vspace{0.1in}

To solve $\textbf{P.3}$, we  introduce problem $\textbf{P.3.1}$ as follows:

 \begin{center}
\framebox[0.7\textwidth][c]{
\enspace
\begin{minipage}[t]{0.7\textwidth}
\small
$\textbf{P.3.1}$
$\max f_{avg}(\pi)$
\textbf{subject to:}  \\
$ |V(\pi, \phi) \cap V_i| \leq \overline{\gamma_i}, \forall i \in [m], \forall \phi \in\{\phi\in O^V: p(\phi)>0\}$.\\
$ \sum_{i\in[m]} \max\{\underline{\gamma_i}, |V(\pi, \phi) \cap V_i|\} \leq c, \forall \phi \in\{\phi\in O^V: p(\phi)>0\}$.
\end{minipage}
}
\end{center}
\vspace{0.1in}

It is easy to verify that  $\textbf{P.3.1}$ is a relaxed problem of $\textbf{P.3}$. Moreover, as discussed earlier, the constraints listed in $\textbf{P.3.1}$ is a matroid constraint \citep{el2020fairness}. Hence, $\textbf{P.3.1}$ is to maximize an adaptive submodular function subject to a matroid constraint. For the monotone case, it has been shown that a simple greedy algorithm achieves a $1/2$ approximation ratio \citep{golovin2011adaptive1}. Most importantly, such greedy algorithm always delivers a feasible solution of $\textbf{P.3}$. When the utility function is non-monotone, there exists a $1/6$ approximation solution to $\textbf{P.3.1}$ \citep{tang2021pointwise}. However, this solution, say $S$, might not be a feasible solution of  $\textbf{P.3.1}$ as there may exist some group, say $V_i$, from which the number of selected items does not meet the lower bound $\underline{\gamma_i}$, that is, $|S\cap V_i|< \underline{\gamma_i}$. For each of such groups, we can randomly pick $\underline{\gamma_i}-|S\cap V_i|$ items from $V_i\setminus S$ as a backup set and add them to $S$ to make it a feasible solution. Let $B$ denote the backup set, noting that the probability that an item from $V_i\setminus S$ is included in $B$ is at most $(\underline{\gamma_i}-|S\cap V_i|)/(k_i-|S\cap V_i|)$ whose value is upper bounded by $\max_{i\in[m]} \underline{\gamma_i}/k_i$. Consider an arbitrary $S$ and its partial realization $\psi_S$, Lemma \ref{lem:g} states  that $\mathbb{E}_{\Phi\succeq \psi_S}[f(S \cup \cdot, \Phi)]$  is a submodular function. Recall that the probability that an item from $V\setminus S$ is included in $B$ is at most $\max_{i\in[m]} \underline{\gamma_i}/k_i$. This, together with Lemma 2.2 in \citep{buchbinder2014submodular}, indicates that $\mathbb{E}_{\Phi\succeq \psi, B}[f(S \cup B, \Phi)]\geq (1-\max_{i\in[m]} \underline{\gamma_i}/k_i)\cdot \mathbb{E}_{\Phi\succeq \psi_S}[f(S, \Phi)]$. This implies that adding a backup set to $S$ leads to a utility reduction of at most  $1-\max_{i\in[m]} \underline{\gamma_i}/k_i$. This, together with the facts that $S$ is a $1/6$ approximation solution to $\textbf{P.3.1}$ and  $\textbf{P.3.1}$ is a relaxed problem of $\textbf{P.3}$, implies that $S \cup B$ is a $(1-\max_{i\in[m]} \underline{\gamma_i}/k_i)/6$-approximation solution to $\textbf{P.3}$.

%\section{Conclusion}
%In this paper, we study the non-adaptive and adaptive submodular maximization problems subject to group equality constraints. We develop a series of constant-factor approximation algorithms for both settings. In the future, we would like to consider other metrics of fairness.
\bibliographystyle{ijocv081}
\bibliography{reference}

\begin{thebibliography}{28}
\expandafter\ifx\csname natexlab\endcsname\relax\def\natexlab#1{#1}\fi
\expandafter\ifx\csname url\endcsname\relax
  \def\url#1{{\tt #1}}\fi
\expandafter\ifx\csname urlprefix\endcsname\relax\def\urlprefix{URL }\fi
\expandafter\ifx\csname urlstyle\endcsname\relax
  \expandafter\ifx\csname doi\endcsname\relax
  \def\doi#1{doi:\discretionary{}{}{}#1}\fi \else
  \expandafter\ifx\csname doi\endcsname\relax
  \def\doi{doi:\discretionary{}{}{}\begingroup \urlstyle{rm}\Url}\fi \fi

\bibitem[{Amanatidis et~al.(2020)Amanatidis, Fusco, Lazos, Leonardi, and
  Reiffenh{\"a}user}]{amanatidis2020fast}
Amanatidis, Georgios, Federico Fusco, Philip Lazos, Stefano Leonardi, Rebecca
  Reiffenh{\"a}user. 2020.
\newblock Fast adaptive non-monotone submodular maximization subject to a
  knapsack constraint.
\newblock {\it Advances in neural information processing systems\/}.

\bibitem[{Bronfenbrenner(1973)}]{bronfenbrenner1973equality}
Bronfenbrenner, Martin. 1973.
\newblock Equality and equity.
\newblock {\it The ANNALS of the American Academy of Political and Social
  Science\/} {\bf 409} 9--23.

\bibitem[{Buchbinder et~al.(2014)Buchbinder, Feldman, Naor, and
  Schwartz}]{buchbinder2014submodular}
Buchbinder, Niv, Moran Feldman, Joseph Naor, Roy Schwartz. 2014.
\newblock Submodular maximization with cardinality constraints.
\newblock {\it Proceedings of the twenty-fifth annual ACM-SIAM symposium on
  Discrete algorithms\/}. SIAM, 1433--1452.

\bibitem[{Calinescu et~al.(2007)Calinescu, Chekuri, P{\'a}l, and
  Vondr{\'a}k}]{calinescu2007maximizing}
Calinescu, Gruia, Chandra Chekuri, Martin P{\'a}l, Jan Vondr{\'a}k. 2007.
\newblock Maximizing a submodular set function subject to a matroid constraint.
\newblock {\it International Conference on Integer Programming and
  Combinatorial Optimization\/}. Springer, 182--196.

\bibitem[{Celis et~al.(2018{\natexlab{a}})Celis, Keswani, Straszak, Deshpande,
  Kathuria, and Vishnoi}]{celis2018fair}
Celis, Elisa, Vijay Keswani, Damian Straszak, Amit Deshpande, Tarun Kathuria,
  Nisheeth Vishnoi. 2018{\natexlab{a}}.
\newblock Fair and diverse dpp-based data summarization.
\newblock {\it International Conference on Machine Learning\/}. PMLR, 716--725.

\bibitem[{Celis et~al.(2018{\natexlab{b}})Celis, Huang, and
  Vishnoi}]{celis2018multiwinner}
Celis, L~Elisa, Lingxiao Huang, Nisheeth~K Vishnoi. 2018{\natexlab{b}}.
\newblock Multiwinner voting with fairness constraints.
\newblock {\it Proceedings of the 27th International Joint Conference on
  Artificial Intelligence\/}. 144--151.

\bibitem[{Das and Kempe(2008)}]{das2008algorithms}
Das, Abhimanyu, David Kempe. 2008.
\newblock Algorithms for subset selection in linear regression.
\newblock {\it Proceedings of the fortieth annual ACM symposium on Theory of
  computing\/}. 45--54.

\bibitem[{Dueck and Frey(2007)}]{dueck2007non}
Dueck, Delbert, Brendan~J Frey. 2007.
\newblock Non-metric affinity propagation for unsupervised image
  categorization.
\newblock {\it 2007 IEEE 11th International Conference on Computer Vision\/}.
  IEEE, 1--8.

\bibitem[{El-Arini and Guestrin(2011)}]{el2011beyond}
El-Arini, Khalid, Carlos Guestrin. 2011.
\newblock Beyond keyword search: discovering relevant scientific literature.
\newblock {\it Proceedings of the 17th ACM SIGKDD international conference on
  Knowledge discovery and data mining\/}. 439--447.

\bibitem[{El~Halabi et~al.(2020)El~Halabi, Mitrovi{\'c}, Norouzi-Fard, Tardos,
  and Tarnawski}]{el2020fairness}
El~Halabi, Marwa, Slobodan Mitrovi{\'c}, Ashkan Norouzi-Fard, Jakab Tardos,
  Jakub~M Tarnawski. 2020.
\newblock Fairness in streaming submodular maximization: algorithms and
  hardness.
\newblock {\it Advances in Neural Information Processing Systems\/} {\bf 33}
  13609--13622.

\bibitem[{Feldman et~al.(2011)Feldman, Naor, and Schwartz}]{feldman2011unified}
Feldman, Moran, Joseph Naor, Roy Schwartz. 2011.
\newblock A unified continuous greedy algorithm for submodular maximization.
\newblock {\it 2011 IEEE 52nd Annual Symposium on Foundations of Computer
  Science\/}. IEEE, 570--579.

\bibitem[{Golovin and Krause(2011{\natexlab{a}})}]{golovin2011adaptive1}
Golovin, Daniel, Andreas Krause. 2011{\natexlab{a}}.
\newblock Adaptive submodular optimization under matroid constraints.
\newblock {\it arXiv preprint arXiv:1101.4450\/} .

\bibitem[{Golovin and Krause(2011{\natexlab{b}})}]{golovin2011adaptive}
Golovin, Daniel, Andreas Krause. 2011{\natexlab{b}}.
\newblock Adaptive submodularity: Theory and applications in active learning
  and stochastic optimization.
\newblock {\it Journal of Artificial Intelligence Research\/} {\bf 42}
  427--486.

\bibitem[{Gotovos et~al.(2015)Gotovos, Karbasi, and Krause}]{gotovos2015non}
Gotovos, Alkis, Amin Karbasi, Andreas Krause. 2015.
\newblock Non-monotone adaptive submodular maximization.
\newblock {\it Twenty-Fourth International Joint Conference on Artificial
  Intelligence\/}.

\bibitem[{Joseph et~al.(2016)Joseph, Kearns, Morgenstern, and
  Roth}]{joseph2016fairness}
Joseph, Matthew, Michael Kearns, Jamie~H Morgenstern, Aaron Roth. 2016.
\newblock Fairness in learning: Classic and contextual bandits.
\newblock {\it Advances in neural information processing systems\/} {\bf 29}.

\bibitem[{Kempe et~al.(2003)Kempe, Kleinberg, and Tardos}]{kempe2003maximizing}
Kempe, David, Jon Kleinberg, {\'E}va Tardos. 2003.
\newblock Maximizing the spread of influence through a social network.
\newblock {\it Proceedings of the ninth ACM SIGKDD international conference on
  Knowledge discovery and data mining\/}. 137--146.

\bibitem[{NSF(2022)}]{nsf}
NSF. 2022.
\newblock Expanding ai innovation through capacity building and partnerships.
\newblock
  \urlprefix\url{https://beta.nsf.gov/funding/opportunities/expanding-ai-innovation-through-capacity-building}.

\bibitem[{Sipos et~al.(2012)Sipos, Swaminathan, Shivaswamy, and
  Joachims}]{sipos2012temporal}
Sipos, Ruben, Adith Swaminathan, Pannaga Shivaswamy, Thorsten Joachims. 2012.
\newblock Temporal corpus summarization using submodular word coverage.
\newblock {\it Proceedings of the 21st ACM international conference on
  Information and knowledge management\/}. 754--763.

\bibitem[{Sun et~al.(2023)Sun, Zhang, Zhang, and Zhang}]{sun2023improved}
Sun, Xiaoming, Jialin Zhang, Shuo Zhang, Zhijie Zhang. 2023.
\newblock Improved deterministic algorithms for non-monotone submodular
  maximization.
\newblock {\it Computing and Combinatorics: 28th International Conference,
  COCOON 2022, Shenzhen, China, October 22--24, 2022, Proceedings\/}. Springer,
  496--507.

\bibitem[{Tang(2021)}]{tang2021beyond}
Tang, Shaojie. 2021.
\newblock Beyond pointwise submodularity: Non-monotone adaptive submodular
  maximization in linear time.
\newblock {\it Theoretical Computer Science\/} {\bf 850} 249--261.

\bibitem[{Tang(2022)}]{tang2021pointwise}
Tang, Shaojie. 2022.
\newblock Beyond pointwise submodularity: Non-monotone adaptive submodular
  maximization subject to knapsack and k-system constraints.
\newblock {\it Theoretical Computer Science\/} {\bf 936} 139--147.
\newblock \doi{https://doi.org/10.1016/j.tcs.2022.09.022}.
\newblock
  \urlprefix\url{https://www.sciencedirect.com/science/article/pii/S0304397522005643}.

\bibitem[{Tang and Yuan(2020)}]{tang2020influence}
Tang, Shaojie, Jing Yuan. 2020.
\newblock Influence maximization with partial feedback.
\newblock {\it Operations Research Letters\/} {\bf 48} 24--28.

\bibitem[{Tang and Yuan(2022)}]{tang2021optimal}
Tang, Shaojie, Jing Yuan. 2022.
\newblock Optimal sampling gaps for adaptive submodular maximization.
\newblock {\it AAAI\/}.

\bibitem[{Tang and Yuan(2023)}]{tang2023beyond}
Tang, Shaojie, Jing Yuan. 2023.
\newblock Beyond submodularity: a unified framework of randomized set selection
  with group fairness constraints.
\newblock {\it Journal of Combinatorial Optimization\/} {\bf 45} 102.

\bibitem[{Tang et~al.(2023)Tang, Yuan, and Twumasi}]{tang2023ctw}
Tang, Shaojie, Jing Yuan, Mensah-Boateng Twumasi. 2023.
\newblock Achieving long-term fairness in submodular maximization through
  randomization.
\newblock {\it 19th Cologne-Twente Workshop on Graphs and Combinatorial
  Optimization\/}.

\bibitem[{Tsang et~al.(2019)Tsang, Wilder, Rice, Tambe, and
  Zick}]{tsang2019group}
Tsang, Alan, Bryan Wilder, Eric Rice, Milind Tambe, Yair Zick. 2019.
\newblock Group-fairness in influence maximization.
\newblock {\it arXiv preprint arXiv:1903.00967\/} .

\bibitem[{Yuan and Tang(2023)}]{yuan2023group}
Yuan, Jing, Shaojie Tang. 2023.
\newblock Group fairness in non-monotone submodular maximization.
\newblock {\it Journal of Combinatorial Optimization\/} {\bf 45} 88.

\bibitem[{Zafar et~al.(2017)Zafar, Valera, Rogriguez, and
  Gummadi}]{zafar2017fairness}
Zafar, Muhammad~Bilal, Isabel Valera, Manuel~Gomez Rogriguez, Krishna~P
  Gummadi. 2017.
\newblock Fairness constraints: Mechanisms for fair classification.
\newblock {\it Artificial intelligence and statistics\/}. PMLR, 962--970.

\end{thebibliography}

\clearpage

\setcounter{page}{1}
\begin{center}Online Supplement\end{center}
\begin{APPENDICES}
\section{Missing Definitions, Lemmas and Proofs}
\subsection{Proof of Lemma \ref{lem:wash}}
\emph{Proof:} Consider an instance of the NP-hard \emph{cardinality constrained submodular maximization problem}, defined by a group of items $U$, a general submodular utility function $h: 2^U\rightarrow \mathbb{R}_+$, and a cardinality constraint $b$; we wish to find a subset of items $S\subseteq U$ to maximize $h(S)$ such that $|S|\leq b$. We next show that this problem is a special case of  $\textbf{P.0}$. Given an arbitrary instance of cardinality constrained submodular maximization problem, we define a corresponding instance of $\textbf{P.0}$ as follows: $V$ is identical to $U$, $f: 2^V\rightarrow \mathbb{R}_+$ is identical to $h: 2^U\rightarrow \mathbb{R}_+$, there are two groups $V_1$ and $V_2$ such that $V_1=V$ and $V_2=\emptyset$, and $\alpha=b$. It is easy to verify that these two instances are equivalent in terms of approximability. This finishes the proof of this lemma. $\Box$

\subsection{Proof of Lemma \ref{lem:g}}
\emph{Proof:} Recall that for any partial realization $\psi$ and any set $S\subseteq V\setminus \mathrm{dom}(\psi)$,
$g_{\psi}(S)=\mathbb{E}_{\Phi\succeq \psi}[f(\mathrm{dom}(\psi)\cup S, \Phi)]= \mathbb{E}_{\Phi\succeq \psi}[f(\mathrm{dom}(\psi), \Phi)]+\Delta(S\mid \psi)$. Because $\mathbb{E}_{\Phi\succeq \psi}[f(\mathrm{dom}(\psi), \Phi)]$ is a constant, to prove this lemma, it suffices to show that $\Delta(\cdot\mid \psi): 2^{V\setminus \mathrm{dom}(\psi)}\rightarrow \mathbb{R}_{\geq0}$ is a submodular function for any $\psi$. The rest of this proof is devoted to proving this.

Let $\Phi(S)=\cup_{e\in S}\Phi(e)$. Consider two subsets $A$ and $B$ such that $A\subseteq B \subseteq V\setminus \mathrm{dom}(\psi)$, and for any $e\in V\setminus (\mathrm{dom}(\psi)\cup B)$,
\begin{eqnarray}
\Delta(A \cup e\mid \psi)- \Delta(A\mid \psi)&=& \mathbb{E}_{\Phi\succeq \psi}[f(\mathrm{dom}(\psi)\cup A\cup \{e\}, \Phi)] - \mathbb{E}_{\Phi\succeq \psi}[f(\mathrm{dom}(\psi)\cup A, \Phi)]\\
&=& \mathbb{E}_{\Phi\succeq \psi}[\Delta(e\mid \psi\cup\Phi(A))]\\
&\geq& \mathbb{E}_{\Phi\succeq \psi}[\Delta(e\mid \psi\cup\Phi(B))]\\
&=& \mathbb{E}_{\Phi\succeq \psi}[f(\mathrm{dom}(\psi)\cup B\cup \{e\}, \Phi)] - \mathbb{E}_{\Phi\succeq \psi}[f(\mathrm{dom}(\psi)\cup B, \Phi)]\\
&=& \Delta(B \cup \{e\}\mid \psi)- \Delta(B\mid \psi).
\end{eqnarray} $\Box$

\subsection{Proof of Lemma \ref{lem:feasible}}
\emph{Proof:} To prove this lemma, it suffices to show that both $A^1$ and  $A^2$ are feasible. We focus on proving that $A^1$ is feasible, and the same argument can be used to prove the feasibility of $A^2$. Recall that for all $i\in L$, where $L=\{i\in[m]\mid |A^{\textsf{greedy}} \cap V_i|< \min\{\lfloor\frac{k_i}{2}\rfloor, \lfloor\frac{k_{\min}}{2}\rfloor+\alpha\}\}$, $X_i$ is a set picked  from  $V_i\setminus A^{\textsf{greedy}}$ such that $|X_i|=\min\{\lfloor\frac{k_i}{2}\rfloor, \lfloor\frac{k_{\min}}{2}\rfloor+\alpha\} - |A^{\textsf{greedy}} \cap V_i|$. Hence, for all $i\in L$, $|(A^{\textsf{greedy}} \cap V_i) \cup X_i|=\min\{\lfloor\frac{k_i}{2}\rfloor, \lfloor\frac{k_{\min}}{2}\rfloor+\alpha\}$. Recall that $A^1=  A^{\textsf{greedy}}\cup\{\cup_{i\in L} X_i\}$, hence, for all $i\in[m]$,
  \begin{eqnarray}
 \label{eq:sharon}
 |A^1 \cap V_i|=\min\{\lfloor\frac{k_i}{2}\rfloor, \lfloor\frac{k_{\min}}{2}\rfloor+\alpha\}.
  \end{eqnarray}

   Moreover, we have that for all $i\in[m]$,
 \begin{eqnarray}
 \label{eq:sharon2}
 \lfloor\frac{k_{\min}}{2}\rfloor \leq \min\{\lfloor\frac{k_i}{2}\rfloor, \lfloor\frac{k_{\min}}{2}\rfloor+\alpha\}\leq \lfloor\frac{k_{\min}}{2}\rfloor+\alpha,
 \end{eqnarray}
 where the first inequality is due to   $k_i\geq k_{\min}$ for all $i\in[m]$. (\ref{eq:sharon}) and (\ref{eq:sharon2}) together imply that $|A^1 \cap V_i|-|A^1 \cap V_j| \leq \alpha, \forall i, j\in[m]$. Hence, $A^1$ is a feasible solution to $\textbf{P.0}$. $\Box$

\subsection{Proof of Lemma \ref{lem:bash}}

\emph{Proof:} Recall that \textsf{Greedy} is a randomized algorithm whose output is dependent on the realization of $R$. For each $e\in V$, let $H(e)$ denote the set that contains all possible runs of \textsf{Greedy} under which $e$ is being considered and it is among \emph{best-looking} items. Let $\mathcal{D}(H(e))$ denote the  prior probability distribution over $H(e)$. In addition, let $H^+(e)$ denote the set of all possible runs of \textsf{Greedy} under which $e$ is being considered and let $\mathcal{D}(H^+(e))$ represent the  prior probability distribution over $H^+(e)$. It is easy to verify that $H(e) \subseteq H^+(e)$.  Consider any item $e\in V$ and any fixed run $\lambda \in H^+(e)$, assume $S^\lambda_{[e]}$ contains all items that are selected before $e$ is being considered and $S^\lambda$ contains all selected items under $\lambda$. Let $\Lambda$ denote a random run of \textsf{Greedy},  we have
\begin{eqnarray}
\label{eq:1112}
  \mathbb{E}[f(A^{\textsf{greedy}})]
     &=& \sum_{e\in V}\mathbb{E}_{\Lambda \sim \mathcal{D}(H^+(e))}[\Pr[e \mbox{ is selected given it was considered}]f(e\mid S^\Lambda_{[e]})]\\
    &\geq& \sum_{e\in V}\mathbb{E}_{\Lambda \sim \mathcal{D}(H(e))}[\Pr[e \mbox{ is selected given it was considered}]f(e\mid S^\Lambda_{[e]})]\\
 &=&  \sum_{e\in V}\mathbb{E}_{\Lambda \sim \mathcal{D}(H(e))}[p\cdot f(e\mid S^\Lambda_{[e]})]\\
  &=& p\cdot  \sum_{e\in V}\mathbb{E}_{\Lambda \sim \mathcal{D}(H(e))}[f(e\mid  S^\Lambda_{[e]})]\\
  &\geq&p\cdot \sum_{e\in V}\mathbb{E}_{\Lambda \sim \mathcal{D}(H(e))}[f(e\mid S^\Lambda)]\\
  &=& p\cdot  \mathbb{E}[\sum_{i\in[m]}\sum_{e\in C_i} f(e\mid S^\Lambda)] = p\cdot  \mathbb{E}[\sum_{i\in[m]}\sum_{e\in C_i} f(e\mid A^{\textsf{greedy}})].
\end{eqnarray}
The first inequality is due to  $H(e) \subseteq H^+(e)$. The second equality is due to for every $e\in V$,  \textsf{Greedy} selects $e$ with probability $p$ given that $e$ has been considered. The second inequality is due to for every $e\in V$ and every $\lambda \in H(e)$, $S^\Lambda_{[e]} \subseteq S^\Lambda$ and $f: 2^V\rightarrow \mathbb{R}_+$ is  submodular. The last equality is due to the assumption that $S^\lambda$ represents the set of items selected by \textsf{Greedy} under a fixed run $\lambda$, hence, $A^{\textsf{greedy}}=S^\lambda$ under $\lambda$. $\Box$

\subsection{Proof of Lemma \ref{lem:summer}}
\emph{Proof:} Recall that $l_i=\min\{|W\cap V_i|, k_i, k_{\min}+\alpha\}$. For each $i\in[m]$, let $r_i=\min\{\lfloor\frac{k_i}{2}\rfloor, \lfloor\frac{k_{\min}}{2}\rfloor+\alpha\}$ denote the size constraint specified in Definition \ref{def:1}. Let  $S^\lambda_i=S^\lambda\cap V_i$ for each $i\in[m]$, we first show that for any fixed run $\lambda$ of \textsf{Greedy} and any group $i\in [m]$, the following inequality holds:
\begin{eqnarray}
\label{eq:white}
\sum_{e\in S^\lambda_i} f(e\mid S^\lambda_{[e]})\geq \frac{r_i}{l_i} \cdot \sum_{e\in U_i}  f(e\mid S^\lambda).
\end{eqnarray}

To prove the above inequality, we consider two scenarios. If $|S^\lambda_i|< r_i$, which implies that the size constraint of group $V_i$ is not binding, then $ f(e\mid S^\lambda)<0$ for all $e\in U_i$ due to the definition of $U_i$ and the design of \textsf{Greedy}. Hence, $ \sum_{e\in U_i}  f(e\mid S^\lambda)<0$ in this scenario. Now consider the case when $|S^\lambda_i|= r_i$. First, because $f: 2^V\rightarrow \mathbb{R}_+$ is submodular and $ S^\lambda \supseteq  S^\lambda_{[e]}$ for all $e\in S^\lambda_i$, we have that for all $e'\in U_i$ and $e\in S^\lambda_i$, $f(e'\mid S^\lambda)\leq f(e'\mid S^\lambda_{[e]})$. Recall that  \textsf{Greedy} selects items in an greedy manner, we have that for all $e'\in U_i$ and $e\in S^\lambda_i$, $f(e'\mid S^\lambda_{[e]}) \leq f(e\mid S^\lambda_{[e]})$. Hence, for all $e'\in U_i$ and $e\in S^\lambda_i$, we have that $f(e'\mid S^\lambda) \leq f(e\mid S^\lambda_{[e]})$. This, together with the assumption that $|S^\lambda_i|= r_i$ and the fact that $|U_i|\leq |D_i| = l_i$, implies that
\begin{eqnarray}
\sum_{e\in S^\lambda_i} f(e\mid S^\lambda_{[e]})\geq \frac{r_i}{l_i} \cdot \sum_{e\in U_i}  f(e\mid S^\lambda).
\end{eqnarray}

Let $\mathcal{D}$ denote the distribution of $\lambda$, it follows that
\begin{eqnarray}
 \mathbb{E}[f(A^{\textsf{greedy}})]
     &=& \mathbb{E}_{\Lambda \sim \mathcal{D}}[ \sum_{e\in S^\Lambda}f(e\mid S^\Lambda_{[e]})] \\
     &=& \mathbb{E}_{\Lambda \sim \mathcal{D}}[\sum_{i\in[m]} \sum_{e\in S^\lambda_i} f(e\mid S^\lambda_{[e]})] \\
     &\geq& \mathbb{E}_{\Lambda \sim \mathcal{D}}[ \sum_{i\in[m]} \frac{r_i}{l_i} \cdot \sum_{e\in U_i}  f(e\mid S^\Lambda)]\\
     &=& \mathbb{E}_{\Lambda \sim \mathcal{D}}[ \sum_{i\in[m]} \frac{\min\{\lfloor\frac{k_i}{2}\rfloor, \lfloor\frac{k_{\min}}{2}\rfloor+\alpha\}}{\min\{|W\cap V_i|, k_i, k_{\min}+\alpha\}} \cdot \sum_{e\in U_i}  f(e\mid S^\Lambda)]\\
         &\geq& \mathbb{E}_{\Lambda \sim \mathcal{D}}[ \sum_{i\in[m]} \frac{\min\{\lfloor\frac{k_i}{2}\rfloor, \lfloor\frac{k_{\min}}{2}\rfloor+\alpha\}}{\min\{ k_i, k_{\min}+\alpha\}} \cdot \sum_{e\in U_i}  f(e\mid S^\Lambda)]\\
                 % &\geq& \mathbb{E}_{\Lambda \sim \mathcal{D}}[ \sum_{i\in[m]} \min_{i\in[m]} \frac{\lfloor k_i/2\rfloor}{k_i}  \times \sum_{e\in U_i} f(e\mid S^\Lambda)]\\
                           &\geq& \mathbb{E}_{\Lambda \sim \mathcal{D}}[ \sum_{i\in[m]} \frac{1}{3} \cdot \sum_{e\in U_i} f(e\mid S^\Lambda)]\label{eq:nainai5}\\
                                  &=& \frac{1}{3}\cdot  \mathbb{E}_{\Lambda \sim \mathcal{D}}[ \sum_{i\in[m]} \sum_{e\in U_i} f(e\mid S^\Lambda)]=  \frac{1}{3}\cdot  \mathbb{E}[\sum_{i\in[m]}\sum_{e\in U_i}f(e \mid A^{\textsf{greedy}})].
\end{eqnarray}
The first inequality is due to (\ref{eq:white}) and the last equality is due to the assumption that $S^\lambda$ represents the set of items selected by \textsf{Greedy} under a fixed run $\lambda$, hence, $A^{\textsf{greedy}}=S^\lambda$ under $\lambda$. To prove the third  inequality, it is sufficient to show that
\begin{eqnarray}
\label{eq:nainai}
\frac{\min\{\lfloor\frac{k_i}{2}\rfloor, \lfloor\frac{k_{\min}}{2}\rfloor+\alpha\}}{\min\{ k_i, k_{\min}+\alpha\}} \geq 1/3
 \end{eqnarray} for all $i\in[m]$. The proof of (\ref{eq:nainai}) is trivial when $\alpha=0$, i.e., if $\alpha=0$, then
 \begin{eqnarray}
\label{eq:nainai2}
&&\frac{\min\{\lfloor\frac{k_i}{2}\rfloor, \lfloor\frac{k_{\min}}{2}\rfloor+\alpha\}}{\min\{ k_i, k_{\min}+\alpha\}} = \frac{\min\{\lfloor\frac{k_i}{2}\rfloor, \lfloor\frac{k_{\min}}{2}\rfloor\}}{\min\{ k_i, k_{\min}\}} \\
&&\geq  \min_{j\in[m]}\frac{\lfloor k_j/2\rfloor}{k_j} \geq 1/3,
 \end{eqnarray}
 where the second inequality is due to the assumption that $k_{\min}> 1$.  We next assume $\alpha>0$ and show that
 \begin{eqnarray}
  \lfloor\frac{k_{\min}}{2}\rfloor+\alpha\geq  \lfloor\frac{k_{\min}+\alpha}{2}\rfloor.
  \end{eqnarray}
 Observe that
   \begin{eqnarray}
  &&\lfloor\frac{k_{\min}}{2}\rfloor+\alpha\geq \frac{k_{\min}}{2}-0.5+\alpha = \frac{k_{\min}}{2}-0.5+\frac{\alpha}{2} + \frac{\alpha}{2}\\
  &&\geq  \frac{k_{\min}+\alpha}{2} \geq \lfloor\frac{k_{\min}+\alpha}{2}\rfloor, \label{eq:nainai3}
  \end{eqnarray}
  where the second inequality is due to the assumption that $\alpha>0$. It follows that
  \begin{eqnarray}
\label{eq:nainai1}
&&\frac{\min\{\lfloor\frac{k_i}{2}\rfloor, \lfloor\frac{k_{\min}}{2}\rfloor+\alpha\}}{\min\{ k_i, k_{\min}+\alpha\}} \geq \frac{\min\{\lfloor\frac{k_i}{2}\rfloor, \lfloor\frac{k_{\min}+\alpha}{2}\rfloor\}}{\min\{ k_i, k_{\min}+\alpha\}} \geq 1/3,
 \end{eqnarray}
 where the first inequality is due to (\ref{eq:nainai3}) and the second inequality is due to the assumption that $k_{\min}> 1$.

  $\Box$

\subsection{Proof of Lemma \ref{lem:sharon-55}}

\emph{Proof:} Lemma \ref{lem:bash} and Lemma \ref{lem:summer} imply that
\begin{eqnarray}
&&(\frac{1}{p}+3)\mathbb{E}[f(A^{\textsf{greedy}})]\geq \mathbb{E}[\sum_{i\in[m]}\sum_{e\in C_i}f(e \mid A^{\textsf{greedy}}) + \sum_{i\in[m]}\sum_{e\in U_i}f(e \mid A^{\textsf{greedy}})]\\
&&\geq f(OPT\mid A^{\textsf{greedy}}),
\end{eqnarray}
where the second inequality is due to (\ref{eq:will}).
It follows that
\begin{eqnarray}
\label{sharon-3}
(\frac{1}{p}+4)\mathbb{E}[f(A^{\textsf{greedy}})]\geq \mathbb{E}[f(A^{\textsf{greedy}})+ f(OPT\mid A^{\textsf{greedy}})]=\mathbb{E}[f(A^{\textsf{greedy}}\cup OPT)].
\end{eqnarray}

Recall that $R$  is a random set that contains each item independently with probability at most $p$, and $A^{\textsf{greedy}}$ is a subset of $R$, hence, $A^{\textsf{greedy}}$ contains each item with probability at most $p$. This, together with Lemma 2.2 in \citep{buchbinder2014submodular}, implies that
\begin{eqnarray}
\label{sharon-2}
\mathbb{E}[f(A^{\textsf{greedy}}\cup OPT)] \geq (1-p) f(OPT).
\end{eqnarray}

(\ref{sharon-2}) and (\ref{sharon-3}) imply that
\begin{eqnarray}
\label{sharon-4}
(\frac{1}{p}+4)\mathbb{E}[f(A^{\textsf{greedy}})] \geq (1-p) f(OPT).
\end{eqnarray}

If we set $p=\frac{\sqrt{5}-1}{4}$, then
\begin{eqnarray}
\mathbb{E}[f(A^{\textsf{greedy}})] \geq 0.09\cdot  f(OPT).
\end{eqnarray}
 $\Box$

 \subsection{Proof of Theorem \ref{thm:1}}

\emph{Proof:} Recall that after obtaining a greedy solution $A^{\textsf{greedy}}$, we construct two candidate solutions  $A^1=  A^{\textsf{greedy}}\cup\{\cup_{i\in L} X_i\}$, $A^2=  A^{\textsf{greedy}}\cup\{\cup_{i\in L} Y_i\}$ such that for each $i\in L$, $X_i\cap Y_i=\emptyset$, which implies that $(\cup_{i\in L} X_i)\cap (\cup_{i\in L} Y_i)=\emptyset$. According to Lemma 1 in \citep{tang2021pointwise}, if $(\cup_{i\in L} X_i)\cap (\cup_{i\in L} Y_i)=\emptyset$ and $f: 2^V\rightarrow \mathbb{R}_+$ is submodular, then $f(A^{\textsf{greedy}}\cup\{\cup_{i\in L} X_i\})+f(A^{\textsf{greedy}}\cup\{\cup_{i\in L} Y_i\}) \geq f(A^{\textsf{greedy}})$. Hence,
$f(A^1)+f(A^2) = f(A^{\textsf{greedy}}\cup\{\cup_{i\in L} X_i\})+f(A^{\textsf{greedy}}\cup\{\cup_{i\in L} Y_i\}) \geq f(A^{\textsf{greedy}})$. Because $A^{\textsf{final}}$ is the better solution between $A^1$ and $A^2$, we have  $f(A^{\textsf{final}}) = \max\{f(A^1), f(A^2)\}\geq \frac{f(A^1)+f(A^2)}{2}\geq \frac{f(A^{\textsf{greedy}})}{2}$. Hence, $\mathbb{E}[f(A^{\textsf{final}})]\geq \mathbb{E}[f(A^{\textsf{greedy}})]/2$. This, together with (\ref{sharon-5}), implies that $\mathbb{E}[f(A^{\textsf{final}})]\geq 0.045 \cdot  f(OPT)$. $\Box$

  \subsection{Proof of Lemma \ref{lem:bash-b}}

 \emph{Proof:}  For each $e\in V$, let $H(e)$ denote the set that contains all possible runs of $\pi^g$ under which $e$ is being considered and it is among \emph{best-looking} items. Let $\mathcal{D}(H(e))$ denote the  prior probability distribution over $H(e)$. In addition, let $H^+(e)$ denote the set of all possible runs of $\pi^g$ under which $e$ is being considered and let $\mathcal{D}(H^+(e))$ represent the  prior probability distribution over $H^+(e)$. It is easy to verify that $H(e) \subseteq H^+(e)$.  Moreover, for each $e\in V$ and $\lambda\in H^+(e)$, let $\psi^\lambda_{[e]}$ denote the partial realization of all selected items before $e$ is being considered under $\lambda$. Then we have
\begin{eqnarray}
\label{eq:1112-b}
f_{avg}(\pi^g)
     &=& \sum_{e\in V}\mathbb{E}_{\Lambda \sim \mathcal{D}(H^+(e))}[\Pr[e \mbox{ is selected given it was considered}]\Delta(e\mid \psi^\Lambda_{[e]})]\\
    &\geq& \sum_{e\in V}\mathbb{E}_{\Lambda \sim \mathcal{D}(H(e))}[\Pr[e \mbox{ is selected given it was considered}]\Delta(e\mid \psi^\Lambda_{[e]})]\\
 &=&  \sum_{e\in V}\mathbb{E}_{\Lambda \sim \mathcal{D}(H(e))}[p\cdot \Delta(e\mid \psi^\Lambda_{[e]})]\\
  &=& p\cdot  \sum_{e\in V}\mathbb{E}_{\Lambda \sim \mathcal{D}(H(e))}[\Delta(e\mid \psi^\Lambda_{[e]})]\\
  &\geq&p\cdot \sum_{e\in V}\mathbb{E}_{\Lambda \sim \mathcal{D}(H(e))}[\Delta(e\mid \psi^\Lambda)]\\
  &=& p\cdot  \mathbb{E}[\sum_{i\in[m]}\sum_{e\in C_i} \Delta(e\mid \psi^\Lambda)].
\end{eqnarray}
The first inequality is due to  $H(e) \subseteq H^+(e)$. The second equality is due to $e$ is selected with probability $p$ given that $e$ has been considered. The second inequality is due to $\psi^\Lambda_{[e]} \subseteq \psi^\Lambda$  and $f : 2^{V\times O} \rightarrow \mathbb{R}_{\geq0}$ is adaptive submodular. $\Box$

  \subsection{Proof of Lemma \ref{lem:summer-b}}

  \emph{Proof:} Recall that $l_i=\min\{|W\cap V_i|, k_i, k_{\min}+\alpha\}$ and $r_i=\min\{\lfloor\frac{k_i}{2}\rfloor, \lfloor\frac{k_{\min}}{2}\rfloor+\alpha\}$. Let $S^\lambda_i= S^\lambda\cap V_i$ for each $i\in [m]$, we first show that for any fixed run $\lambda$ of $\pi^g$ and any group $i\in [m]$, the following inequality holds:
\begin{eqnarray}
\label{eq:kaison}
\sum_{e\in S^\lambda_i} \Delta(e\mid \psi^\lambda_{[e]})\geq \frac{r_i}{l_i} \cdot \sum_{e\in U_i}  \Delta(e\mid \psi^\lambda).
\end{eqnarray}

To prove the above inequality, we consider two scenarios. If $|S^\lambda_i|< r_i$, then $ \Delta(e\mid \psi^\lambda)<0$ for all $e\in U_i$ due to the definition of $U_i$ and the design of $\pi^g$. Hence, $ \sum_{e\in U_i}  \Delta(e\mid \psi^\lambda)<0$ in this scenario. Now consider the case when $|S^\lambda_i|= r_i$. First, because $f : 2^{V\times O} \rightarrow \mathbb{R}_{\geq0}$ is adaptive submodular and $ \psi^\lambda_i \supseteq  \psi^\lambda_{[e]}$ for all $e\in S^\lambda_i$, we have that for all $e'\in U_i$ and $e\in S^\lambda_i$, $\Delta(e'\mid \psi^\lambda)\leq \Delta(e\mid \psi^\lambda_{[e]})$. Recall that  $\pi^g$ selects items in an greedy manner, we have that for all $e'\in U_i$ and $e\in S^\lambda_i$, $\Delta(e'\mid \psi^\lambda_{[e]}) \leq \Delta(e\mid S^\lambda_{[e]})$. Hence, for all $e'\in U_i$ and $e\in S^\lambda_i$, we have  $\Delta(e'\mid \psi^\lambda) \leq \Delta(e\mid \psi^\lambda_{[e]})$. This, together with the assumption that $|S^\lambda_i|= r_i$ and the fact that $|U_i|\leq |W\cap V_i|= l_i$, implies that
\begin{eqnarray}
\sum_{e\in S^\lambda_i} \Delta(e\mid \psi^\lambda_{[e]})\geq \frac{r_i}{l_i} \cdot \sum_{e\in U_i}  \Delta(e\mid \psi^\lambda).
\end{eqnarray}

It follows that
\begin{eqnarray}
f_{avg}(\pi^g)
     &=& \mathbb{E}_{\Lambda \sim \mathcal{D}}[ \sum_{e\in S^\Lambda}\Delta(e\mid \psi^\Lambda_{[e]})] \\
     &=& \mathbb{E}_{\Lambda \sim \mathcal{D}}[\sum_{i\in[m]} \sum_{e\in S^\lambda_i} \Delta(e\mid \psi^\lambda_{[e]})] \\
     &\geq& \mathbb{E}_{\Lambda \sim \mathcal{D}}[ \sum_{i\in[m]} \frac{r_i}{l_i} \cdot \sum_{e\in U_i}  \Delta(e\mid \psi^\Lambda)]\\
     &=& \mathbb{E}_{\Lambda \sim \mathcal{D}}[ \sum_{i\in[m]} \frac{\min\{\lfloor\frac{k_i}{2}\rfloor, \lfloor\frac{k_{\min}}{2}\rfloor+\alpha\}}{\min\{|W\cap V_i|, k_i, k_{\min}+\alpha\}} \cdot \sum_{e\in U_i}  \Delta(e\mid \psi^\Lambda)]\\
         &\geq& \mathbb{E}_{\Lambda \sim \mathcal{D}}[ \sum_{i\in[m]} \frac{\min\{\lfloor\frac{k_i}{2}\rfloor, \lfloor\frac{k_{\min}}{2}\rfloor+\alpha\}}{\min\{ k_i, k_{\min}+\alpha\}} \cdot \sum_{e\in U_i}  \Delta(e\mid \psi^\Lambda)]\\
                 % &\geq& \mathbb{E}_{\Lambda \sim \mathcal{D}}[ \sum_{i\in[m]} \min_{i\in[m]} \frac{\lfloor k_i/2\rfloor}{k_i}  \times \sum_{e\in U_i} \Delta(e\mid S^\Lambda)]\\
                           &\geq& \mathbb{E}_{\Lambda \sim \mathcal{D}}[ \sum_{i\in[m]} \frac{1}{3} \cdot \sum_{e\in U_i} \Delta(e\mid \psi^\Lambda)]\\
                                  &=& \frac{1}{3} \cdot \mathbb{E}_{\Lambda \sim \mathcal{D}}[ \sum_{i\in[m]} \sum_{e\in U_i} \Delta(e\mid \psi^\Lambda)].
\end{eqnarray}
The first inequality is due to (\ref{eq:kaison}) and the third inequality is by the same proof of (\ref{eq:nainai5}). $\Box$

\subsection{Proof of Lemma \ref{lem:sharon-55-b}}
  \emph{Proof:} Lemma \ref{lem:bash-b} and Lemma \ref{lem:summer-b} imply that
\begin{eqnarray}
&&(\frac{1}{p}+3)f_{avg}(\pi^g) \geq \mathbb{E}_{\Lambda}[\sum_{i\in[m]}\sum_{e\in C_i}\Delta(e \mid \psi^\Lambda)+\sum_{i\in[m]}\sum_{e\in U_i}\Delta(e \mid \psi^\Lambda)] \\
&&\geq \mathbb{E}_{\Lambda}[\Delta(\pi^* \mid \psi^\Lambda)],
\end{eqnarray}
where the second inequality is due to (\ref{eq:will-b}).
It follows that
\begin{eqnarray}
\label{sharon-3-b}
(\frac{1}{p}+4)f_{avg}(\pi^g) \geq f_{avg}(\pi^g)+\mathbb{E}_{\Lambda}[\Delta(\pi^* \mid \psi^\Lambda)]=f_{avg}(\pi^g\cup \pi^*).
\end{eqnarray}

Recall that in the original implementation of $\pi^g$ (Algorithm \ref{alg:2}), $R$  is a random set that contains each item independently with probability at most $p$, and $A^{\textsf{a-greedy}}$ is a subset of $R$. Hence, $A^{\textsf{a-greedy}}$ contains each item with probability at most $p$. Lemma 1 in \citep{tang2021pointwise} shows that if we set $p=1/2$, then
\begin{eqnarray}
\label{sharon-2-b}
f_{avg}(\pi^g\cup \pi^*) \geq f_{avg}(\pi^*) /2.
\end{eqnarray}

(\ref{sharon-2-b}) and (\ref{sharon-3-b}) imply that if we set $p=1/2$,
\begin{eqnarray}
\label{sharon-4-b}
(\frac{1}{p}+4)f_{avg}(\pi^g) = 6 \cdot f_{avg}(\pi^g) \geq f_{avg}(\pi^g\cup \pi^*) \geq f_{avg}(\pi^*) /2.
\end{eqnarray}

Hence,
\begin{eqnarray}
f_{avg}(\pi^g) \geq f_{avg}(\pi^*) /12.
\end{eqnarray} $\Box$

\subsection{Proof of Theorem \ref{thm:1-b}}

\emph{Proof:} Recall that after obtaining $A^{\textsf{a-greedy}}$ from $\pi^g$, we construct two candidate solutions  $A^1=  A^{\textsf{a-greedy}}\cup\{\cup_{i\in L} X_i\}$, $A^2=  A^{\textsf{a-greedy}}\cup\{\cup_{i\in L} Y_i\}$ such that for each $i\in L$, $X_i\cap Y_i=\emptyset$, which implies that $(\cup_{i\in L} X_i)\cap (\cup_{i\in L} Y_i)=\emptyset$. Moreover, in Lemma \ref{lem:g}, we show that for any partial realization $\psi^\lambda$, $g_{\psi^\lambda} : 2^{V\setminus \mathrm{dom}(\psi)}\rightarrow \mathbb{R}_{\geq0} $ is a submodular function. According to Lemma 1 in \citep{tang2021pointwise}, if $(\cup_{i\in L} X_i)\cap (\cup_{i\in L} Y_i)=\emptyset$ and $g_{\psi^\lambda} : 2^{V\setminus \mathrm{dom}(\psi)} \rightarrow \mathbb{R}_{\geq0}$ is submodular, then $g_{\psi^\lambda}(\cup_{i\in L} X_i)+g_{\psi^\lambda}(\cup_{i\in L} Y_i)\geq g_{\psi^\lambda}(\emptyset)$. Because $\pi^f$ selects the better solution between $\mathrm{dom}(\psi^\lambda)\cup (\cup_{i\in L} X_i)$ and $\mathrm{dom}(\psi^\lambda)\cup (\cup_{i\in L} Y_i)$ as the final solution, its utility is at least $g_{\psi^\lambda}(\emptyset)/2$. It follows that $f_{avg}(\pi^f)\geq \mathbb{E}_{\Lambda\sim \mathcal{D}}[g_{\psi^\Lambda}(\emptyset)/2]=\mathbb{E}_{\Lambda\sim \mathcal{D}, \Phi\sim \psi^\Lambda }[f(\mathrm{dom}(\psi^\Lambda), \Phi)/2]=f_{avg}(\pi^g)/2$, where the first equality is due to the definition of  $g_{\psi^\lambda} : 2^{V\setminus \mathrm{dom}(\psi)}\rightarrow \mathbb{R}_{\geq0} $. This, together with Lemma \ref{lem:sharon-55-b}, implies that $f_{avg}(\pi^f) \geq f_{avg}(\pi^*)/24$. $\Box$

\subsection{Proof of Lemma \ref{lem:summer-b-1}}
\emph{Proof:} Recall that $l_i=\min\{|W\cap V_i|, k_i, \alpha+1\}$ and let $r_i=\min\{k_i, \alpha\}$. Let $S^\lambda_i= S^\lambda\cap V_i$ for each $i\in [m]$, we first show that for any fixed run $\lambda$ of $\pi^{f1}$ and any group $i\in [m]$, the following inequality holds:
\begin{eqnarray}
\label{eq:kaison-1}
\sum_{e\in S^\lambda_i} \Delta(e\mid \psi^\lambda_{[e]})\geq \frac{r_i}{l_i} \cdot \sum_{e\in U_i}  \Delta(e\mid \psi^\lambda).
\end{eqnarray}

We consider two scenarios depending on the relation between  $|S^\lambda_i|$ and $r_i$. If $|S^\lambda_i|< r_i$, then $ \Delta(e\mid \psi^\lambda)<0$ for all $e\in U_i$ due to the definition of $U_i$ and the design of $\pi^{f1}$. Hence, $ \sum_{e\in U_i}  \Delta(e\mid \psi^\lambda)<0$ in this scenario. Now consider the case when $|S^\lambda_i|= r_i$. First, because $f : 2^{V\times O} \rightarrow \mathbb{R}_{\geq0}$ is adaptive submodular and $ \psi^\lambda_i \supseteq  \psi^\lambda_{[e]}$ for all $e\in S^\lambda_i$, we have that for all $e'\in U_i$ and $e\in S^\lambda_i$, $\Delta(e'\mid \psi^\lambda)\leq \Delta(e\mid \psi^\lambda_{[e]})$. Recall that  $\pi^{f1}$ selects items in an greedy manner, we have that for all $e'\in U_i$ and $e\in S^\lambda_i$, $\Delta(e'\mid \psi^\lambda_{[e]}) \leq \Delta(e\mid S^\lambda_{[e]})$. Hence, for all $e'\in U_i$ and $e\in S^\lambda_i$, we have  $\Delta(e'\mid \psi^\lambda) \leq \Delta(e\mid \psi^\lambda_{[e]})$. This, together with the assumption that $|S^\lambda_i|= r_i$ and the fact that $|U_i|\leq |W\cap V_i|= l_i$, implies that
\begin{eqnarray}
\sum_{e\in S^\lambda_i} \Delta(e\mid \psi^\lambda_{[e]})\geq \frac{r_i}{l_i} \cdot \sum_{e\in U_i}  \Delta(e\mid \psi^\lambda).
\end{eqnarray}

It follows that
\begin{eqnarray}
f_{avg}(\pi^g)
     &=& \mathbb{E}_{\Lambda \sim \mathcal{D}}[ \sum_{e\in S^\Lambda}\Delta(e\mid \psi^\Lambda_{[e]})] \\
     &=& \mathbb{E}_{\Lambda \sim \mathcal{D}}[\sum_{i\in[m]} \sum_{e\in S^\lambda_i} \Delta(e\mid \psi^\lambda_{[e]})] \\
     &\geq& \mathbb{E}_{\Lambda \sim \mathcal{D}}[ \sum_{i\in[m]} \frac{r_i}{l_i} \cdot \sum_{e\in U_i}  \Delta(e\mid \psi^\Lambda)]\\
     &=& \mathbb{E}_{\Lambda \sim \mathcal{D}}[ \sum_{i\in[m]} \frac{\min\{k_i, \alpha\}}{\min\{|W\cap V_i|, k_i, \alpha+1\}} \cdot \sum_{e\in U_i}  \Delta(e\mid \psi^\Lambda)]\\
         &\geq& \mathbb{E}_{\Lambda \sim \mathcal{D}}[ \sum_{i\in[m]} \frac{\min\{k_i, \alpha\}}{\min\{k_i, \alpha+1\}} \cdot \sum_{e\in U_i}  \Delta(e\mid \psi^\Lambda)]\\
                 % &\geq& \mathbb{E}_{\Lambda \sim \mathcal{D}}[ \sum_{i\in[m]} \min_{i\in[m]} \frac{\lfloor k_i/2\rfloor}{k_i}  \times \sum_{e\in U_i} \Delta(e\mid S^\Lambda)]\\
                           &\geq& \mathbb{E}_{\Lambda \sim \mathcal{D}}[ \sum_{i\in[m]} \frac{1}{2} \cdot \sum_{e\in U_i} \Delta(e\mid \psi^\Lambda)]\\
                                  &=& \frac{1}{2} \cdot \mathbb{E}_{\Lambda \sim \mathcal{D}}[ \sum_{i\in[m]} \sum_{e\in U_i} \Delta(e\mid \psi^\Lambda)].
\end{eqnarray}
The first inequality is due to (\ref{eq:kaison-1}). To prove the third inequality, it is sufficient to show that
\begin{eqnarray}
\frac{\min\{k_i, \alpha\}}{\min\{k_i, \alpha+1\}}\geq 1/2
\end{eqnarray} for all $i\in [m]$. Recall that $k_{\min}=1$ and $\alpha\geq 1$ by our assumptions. We consider two cases depending on the relation between $k_i$ and $\alpha$. If $k_i\leq \alpha$, then $\frac{\min\{k_i, \alpha\}}{\min\{k_i, \alpha+1\}} = \frac{k_i}{k_i} \geq 1/2$. If $k_i\geq \alpha+1$, then $\frac{\min\{k_i, \alpha\}}{\min\{k_i, \alpha+1\}} = \frac{\alpha}{\alpha+1} \geq 1/2$ since $\alpha \geq 1$. $\Box$

\subsection{Proof of Theorem \ref{lem:sharon-55-b-1}}

\emph{Proof:} Lemma \ref{lem:bash-b-1} and Lemma \ref{lem:summer-b-1} imply that
\begin{eqnarray}
&&(\frac{1}{p}+2)f_{avg}(\pi^{f1}) \geq \mathbb{E}_{\Lambda}[\sum_{i\in[m]}\sum_{e\in C_i}\Delta(e \mid \psi^\Lambda)+\sum_{i\in[m]}\sum_{e\in U_i}\Delta(e \mid \psi^\Lambda)] \\
&&\geq \mathbb{E}_{\Lambda}[\Delta(\pi^* \mid \psi^\Lambda)],
\end{eqnarray}
where the second inequality is due to (\ref{eq:will-b-1}).
It follows that
\begin{eqnarray}
\label{sharon-3-b-1}
(\frac{1}{p}+3)f_{avg}(\pi^{f1}) \geq f_{avg}(\pi^{f1})+\mathbb{E}_{\Lambda}[\Delta(\pi^* \mid \psi^\Lambda)]=f_{avg}(\pi^{f1}\cup \pi^*).
\end{eqnarray}

Recall that  $A^{\textsf{final}}$ contains each item with probability at most $p$. According to Lemma 1 in \citep{tang2021pointwise}, if we set $p=1/2$, then
\begin{eqnarray}
\label{sharon-2-b-1}
f_{avg}(\pi^{f1}\cup \pi^*) \geq f_{avg}(\pi^*) /2.
\end{eqnarray}

(\ref{sharon-2-b-1}) and (\ref{sharon-3-b-1}) imply that if we set $p=1/2$, then
\begin{eqnarray}
\label{sharon-4-b-1}
(\frac{1}{p}+3)f_{avg}(\pi^{f1}) = 5 \cdot f_{avg}(\pi^{f1}) \geq f_{avg}(\pi^{f1}\cup \pi^*) \geq f_{avg}(\pi^*) /2.
\end{eqnarray}

Hence,
\begin{eqnarray}
f_{avg}(\pi^{f1}) \geq f_{avg}(\pi^*) /10.
\end{eqnarray} $\Box$

\subsection{Proof of the existence of $M$}
\label{sec:proof-of-m-app}
To prove the existence of such a $M$, we assume that the optimal solution $OPT$ is given.  Then we build $M$ through picking a subset of items from $OPT$ using a greedy algorithm. The greedy algorithm starts with an empty set $M=\emptyset$. In each subsequent iteration, it finds an item with the largest marginal gain from $OPT$ such that adding that item to $M$ does not violate the following condition: For each $i\in[m]$, $|M\cap V_i|\leq \lfloor\frac{|OPT_i|}{2}\rfloor$. This process iterates until $M$ can not be further expanded.
 It is easy to verify that during the implementation of the greedy algorithm, the largest marginal gain cannot be negative. We can prove this by contradiction. Let us assume that $f(e\mid M_t)<0$ for some item $e\in OPT\setminus M_t$ and some intermediate solution set $M_t\subseteq OPT\setminus{e}$. Since $f$ is submodular, we have $f(e\mid M_t)\geq f(e\mid OPT\setminus \{e\})$. This, along with the assumption that $f(e\mid M_t)<0$, implies that $f(e\mid OPT\setminus \{e\})< 0$. Hence, we can remove $e$ from $OPT$ to obtain a better solution, which contradicts the assumption that $OPT$ is the optimal solution.
Therefore, in the implementation of the greedy algorithm, the incremental benefit of adding any item to the current solution set (hence the largest marginal gain) is always non-negative.

We next show that the $M$ returned from the above greedy algorithm satisfies all the aforementioned three conditions. For simplicity, we define $OPT_i=OPT\cap V_i$ for each $i\in[m]$. First, it is easy to verify that when the greedy algorithm terminates, we must have that $\forall i\in[m]$, $|M\cap V_i|=\lfloor\frac{|OPT_i|}{2}\rfloor$. Hence, condition 2 is satisfied. We next prove that $f(M)\geq \frac{\kappa}{1+\kappa}  f(OPT)$ (condition 1). To prove this, we will use the submodularity of $f: 2^V\rightarrow \mathbb{R}_+$. For each $e\in M$, let $M(e)$ denote the partial solution before $e$ is being  selected. Hence, $f(M)=\sum_{e\in M} f(e\mid M(e))$. It follows that
\begin{eqnarray}
f(OPT) &\leq& f(M) + \sum_{e\in OPT\setminus M} f(e \mid M)\\
&=&\sum_{i\in [m]} \sum_{e\in M\cap V_i} f(e\mid M(e))+ \sum_{i\in [m]} \sum_{e\in (OPT\setminus M)\cap V_i} f(e \mid M)\\
&=& \sum_{i\in [m]}( \sum_{e\in M\cap V_i} f(e\mid M(e))+  \sum_{e\in (OPT\setminus M)\cap V_i} f(e \mid M))\\
&\leq& \sum_{i\in [m]}( \sum_{e\in M\cap V_i} f(e\mid M(e))+  \frac{1}{\kappa}\sum_{e\in M\cap V_i} f(e\mid M(e)))\label{eq:min}\\
&=& \sum_{i\in [m]}( (1+\frac{1}{\kappa}) \sum_{e\in M\cap V_i} f(e\mid M(e)))\\
&=& (1+\frac{1}{\kappa})f(M).
\end{eqnarray}

The first inequality is due to $f: 2^V\rightarrow \mathbb{R}_+$ is submodular and the second inequality is due to the following observation: Recall that the greedy algorithm always picks the item with the largest marginal utility in each round. Hence, for each $e\in M$ and $e'\in OPT\setminus M$, we have $f(e' \mid M(e))\leq f(e \mid M(e))$. This, together with the facts that $M(e)\subseteq M$ and  $f: 2^V\rightarrow \mathbb{R}_+$  is submodular, implies that for each $e\in M$ and $e'\in OPT\setminus M$, we have $f(e' \mid M(e))\leq f(e' \mid M)\leq f(e \mid M(e))$.  It follows that for each $i\in[m]$, $\max_{e\in (OPT\setminus M)\cap V_i}f(e \mid M)\leq \min_{e\in M \cap V_i}f(e \mid M(e))$. This, together with $\kappa=\min_{i\in [m]}\frac{|M\cap V_i|}{|OPT\cap V_i|}$, implies (\ref{eq:min}).

At last, we focus on proving condition 3. The proof of $|M|\leq c$ is trivial. Because for each $i\in[m]$, $|M\cap V_i|=\lfloor\frac{|OPT_i|}{2}\rfloor$, we have $|M|\leq |OPT|\leq c$, where the second inequality is due to $OPT$ is a feasible solution. We next prove the first part of condition 3. We first consider the case when $\alpha=0$, i.e., for each $i,j \in[m]$, $|OPT_i|-|OPT_j|= 0$. The proof for this case is trivial because for each $i,j \in[m]$, $|M \cap V_i|-|M \cap V_j|=\lfloor\frac{|OPT_i|}{2}\rfloor-\lfloor\frac{|OPT_j|}{2}\rfloor= 0$.  We next assume that $\alpha\geq1$. Because $OPT$ is feasible, we have that for each $i,j \in[m]$, $|OPT_i|-|OPT_j|\leq \alpha$. It follows that  for each $i,j \in[m]$, $\lfloor\frac{|OPT_i|}{2}\rfloor-\lfloor\frac{|OPT_j|}{2}\rfloor\leq \frac{|OPT_i|}{2}- (\frac{|OPT_j|}{2}-0.5)=\frac{|OPT_i|}{2}- \frac{|OPT_j|}{2}+0.5\leq \frac{\alpha}{2}+0.5 \leq \alpha$, where the second inequality is due to the assumption that $\alpha\geq1$. It follows that for each $i,j \in[m]$, $|M \cap V_i|-|M \cap V_j|\leq \alpha$. Hence, if we let $z=\min_{i\in[m]}|M\cap V_i|$, then  for each $i\in[m]$, $z \leq |M\cap V_i|\leq z+\alpha$.

\subsection{Proof of Lemma \ref{lem:hospital1}}
\emph{Proof:} The proof of the first part is trivial. Consider any feasible solution $S$ to $\textbf{P.2.1}$, observe that for each $i\in[m]$, $z \leq |S\cap V_i|\leq z+\alpha$ implies that $|S \cap V_i|-|S \cap V_j| \leq \alpha, \forall i, j\in[m]$. Hence, $S$ satisfies the group equality constraint. Meanwhile, $S$ also satisfies the cardinality constraint, i.e., $|S|\leq c$. Thus, $S$ is a feasible solution to $\textbf{P.2}$. We next focus on proving that $f(S^{P21})\geq \frac{\kappa}{1+\kappa} f(OPT)$. Using the fact that $M$ satisfies all three conditions listed in the previous section immediately concludes that $M$ is a feasible solution to $\textbf{P.2.1}$ and $f(M)\geq \frac{\kappa}{1+\kappa} f(OPT)$. Because  $S^{P21}$ is the optimal solution to $\textbf{P.2.1}$, we have $f(S^{P21})\geq f(M)\geq \frac{\kappa}{1+\kappa} f(OPT)$. This finishes the proof of this lemma. $\Box$

\subsection{Proof of Lemma \ref{lem:jiaoshi}}
\emph{Proof:} To prove this lemma, it suffices to show that both $A^1$ and  $A^2$ are feasible to $\textbf{P.2.1}$. We focus on proving that $A^1$ is feasible, and the same argument can be used to prove that $A^2$ is feasible. Recall that for all $i\in L$, where $L=\{i\in[m]\mid |A^{P22} \cap V_i|< z\}$, $X_i$ is a set picked  from    $V_i\setminus A^{P22}$ such that $|X_i|=z-|A^{P22} \cap V_i|$. Hence, for all $i\in L$, $|A^1 \cap V_i|=|(A^{P22}\cup X_i) \cap V_i|= z$. By the definition of $L$, we have for all $i\in [m]$, $|A^1 \cap V_i|\geq z$. This finishes the proof of the lower bound. To prove the upper bound, observe that $A^{P22}$ is a feasible solution to $\textbf{P.2.2}$, hence, for each $i\in[m]\setminus L$, we have $|A^{P22} \cap V_i|\leq z+\alpha$. This, together with the fact that for all $i\in L$, $|A^1 \cap V_i|=|(A^{P22}\cup X_i) \cap V_i|= z$, implies that $|A^1 \cap V_i|\leq z+\alpha$ for all $i\in[m]$. This finishes the proof of the upper bound. At last, because $A^{P22}$ is a feasible solution to $\textbf{P.2.2}$, we have $\sum_{i\in[m]}\max\{z, |A^{P22} \cap V_i|\} \leq c$. Meanwhile, because for all $i\in L$, $|A^1 \cap V_i|= z$ and for all $i\in[m]\setminus L$, $|A^1 \cap V_i|= |A^{P22} \cap V_i|\geq z$, where the inequality is due to the definition of $L$, we have $\sum_{i\in[m]} |A^1 \cap V_i|=\sum_{i\in[m]}\max\{z, |A^{P22} \cap V_i|\}$. It follows that $\sum_{i\in[m]} |A^1 \cap V_i|\leq c$. This finishes the proof of the global cardinality constraint. $\Box$

\subsection{Proof of Theorem \ref{lem:yao}}

\emph{Proof:} First, Lemma \ref{lem:hospital1} and Lemma \ref{lem:hospital} imply that $f(S^{P22})\geq f(S^{P21})\geq \frac{\kappa}{1+\kappa}  f(OPT)$, where $S^{P21}$ is the optimal solution to $\textbf{P.2.1}$ and $S^{P22}$ denotes the optimal solution to $\textbf{P.2.2}$. This, together with the fact that $\mathbb{E}[f(A^{P22})]\geq (\frac{1}{e}-o(1))f(S^{P22})$, implies that
\begin{eqnarray}
\label{eq:baylor}\mathbb{E}[f(A^{P22})]\geq (\frac{1}{e}-o(1)) \frac{\kappa}{1+\kappa}  f(OPT).
\end{eqnarray}

Recall that $A^1=  A^{P22}\cup(\cup_{i\in L} X_i)$, $A^2=  A^{P22}\cup(\cup_{i\in L} Y_i)$, and for each $i\in L$, $X_i\cap Y_i=\emptyset$, which implies that $(\cup_{i\in L} X_i)\cap (\cup_{i\in L} Y_i)=\emptyset$. According to Lemma 1 in \citep{tang2021pointwise}, if $(\cup_{i\in L} X_i)\cap (\cup_{i\in L} Y_i)=\emptyset$ and $f: 2^V\rightarrow \mathbb{R}_+$ is submodular, then $f(A^{P22}\cup(\cup_{i\in L} X_i))+f(A^{P22}\cup(\cup_{i\in L} Y_i)) \geq f(A^{P22})$. Hence,
$f(A^1)+f(A^2) = f(A^{P22}\cup(\cup_{i\in L} X_i))+f(A^{P22}\cup(\cup_{i\in L} Y_i)) \geq f(A^{P22})$. Because $A^{\textsf{final}}$ is the better solution between $A^1$ and $A^2$, we have $f(A^{\textsf{final}}) = \max\{f(A^1), f(A^2)\}\geq \frac{f(A^1)+f(A^2)}{2}\geq \frac{f(A^{P22})}{2}$. This, together with (\ref{eq:baylor}), implies that $\mathbb{E}[f(A^{\textsf{final}})]\geq (\frac{1}{e}-o(1))\frac{\kappa}{2(1+\kappa)}  f(OPT)$. Recall that $\kappa=\min_{i\in [m]}\frac{|M\cap V_i|}{|OPT\cap V_i|}=\min_{i\in [m]}\frac{\lfloor\frac{|OPT_i|}{2}\rfloor}{|OPT_i|}$. If $\min_{i\in [m]}|OPT_i|> 1$, then $\kappa=\min_{i\in [m]}\frac{\lfloor\frac{|OPT_i|}{2}\rfloor}{|OPT_i|}\geq 1/3$. Hence, $\mathbb{E}[f(A^{\textsf{final}})]\geq (\frac{1}{e}-o(1))\frac{\kappa}{2(1+\kappa)}  f(OPT) \geq \frac{1/e-o(1)}{8}\cdot f(OPT)$. $\Box$

\subsection{Solving the Case when $k_{\min} \leq 1$}
\label{sec:non-adaptive-special-app}
So far we assume that $k_{\min}> 1$,  now we are ready to tackle the case when $k_{\min}\leq 1$. In this case, $k_{\min}$ has two possible values: $0$ or $1$.
\subsubsection{$k_{\min}=0$}
\label{sec:non-adaptive-special1}
The case when $k_{\min}=0$, i.e., there exists some empty group, is trivial. It is easy to verify that to satisfy the group equality constraint, we must have that for every feasible solution $S$, $|S \cap V_i| \leq \alpha, \forall i\in[m]$. Hence, when $k_{\min}=0$,  our problem is reduced to a classic submodular maximization problem subject to a matroid constraint. We can apply the state-of-the-art algorithm in  \citep{feldman2011unified} to achieve an approximation ratio of   $\frac{1}{e}-o(1)$.
\subsubsection{$k_{\min}=1$}
\label{sec:non-adaptive-special2}
Next we focus on the case when $k_{\min}=1$, i.e., the smallest group contains exactly one item. We consider two subcases: $\min_{i\in[m]}|OPT_i|=0$ and $\min_{i\in[m]}|OPT_i|=1$. Although we do not have the knowledge about $\min_{i\in[m]}|OPT_i|$ initially, we can guess its value, for each guess, we solve the problem to obtain a candidate solution. Finally, the best solution is returned as the final output. The rest of this section is devoted to developing approximation algorithms for each guess.

The case when $\min_{i\in[m]}|OPT_i|=0$, i.e., the optimal solution does not select any items from some group, is trivial. Using the same argument that is used to tackle the case when $k_{\min}=0$, we can convert our problem to a classic submodular maximization problem subject to a matroid constraint.

Now we are left to handle the case when $\min_{i\in[m]}|OPT_i|=1$. This, together with the assumption that $k_{\min}=1$, implies that the optimal solution must select all items from those groups whose size is one. Let $T=\{i\in[m]\mid |V_i|=1\}$ denote the set of the indexes of those groups whose size is one. Because $OPT$ select all items from $\cup_{i\in T} V_i$, it is safe to add $\cup_{i\in T} V_i$ to our solution in advance, leading to an optimization problem listed in $\textbf{P.0.1}$. The objective of  $\textbf{P.0.1}$ is $f'(\cdot)=f(\cdot\cup (\cup_{i\in T} V_i))$, which is a submodular function.

 \begin{center}
\framebox[0.7\textwidth][c]{
\enspace
\begin{minipage}[t]{0.7\textwidth}
\small
$\textbf{P.0.1}$
$\max f'(S)$ \\
\textbf{subject to:}
$S\subseteq V\setminus \cup_{i\in T} V_i$ and
$1 \leq |S \cap V_i| \leq \alpha+1, \forall i \in [m]\setminus T$.
\end{minipage}
}
\end{center}
\vspace{0.1in}

Let $S^{P01}$ denote the optimal solution to $\textbf{P.0.1}$. It is easy to verify that $f'(S^{P01})\geq f'(OPT\setminus \cup_{i\in T} V_i)$, this is because $OPT\setminus \cup_{i\in T} V_i$ is a feasible solution to $\textbf{P.0.1}$. By the definition of $f': 2^{V\setminus \cup_{i\in T} V_i}\rightarrow \mathbb{R}_+$, we have
\begin{eqnarray}
\label{eq:hot}
f(S^{P01}\cup_{i\in T} V_i)\geq f(OPT).
 \end{eqnarray}
Hence, any approximate solution to $\textbf{P.0.1}$ immediately implies an approximate solution to our original problem $\textbf{P.0}$. We next focus on solving $\textbf{P.0.1}$. To this end, we introduce a relaxation of $\textbf{P.0.1}$ as follows.

 \begin{center}
\framebox[0.7\textwidth][c]{
\enspace
\begin{minipage}[t]{0.7\textwidth}
\small
$\textbf{P.0.2}$
$\max f'(S)$ \\
\textbf{subject to:}
$S\subseteq V\setminus \cup_{i\in T} V_i$ and
$|S \cap V_i| \leq \alpha+1, \forall i \in [m]\setminus T$.
\end{minipage}
}
\end{center}
\vspace{0.1in}

Because $f': 2^{V\setminus \cup_{i\in T} V_i}\rightarrow \mathbb{R}_+$ is a submodular function, $\textbf{P.0.2}$ is a classic submodular maximization problem subject to a matroid constraint. Again, we can apply the randomized algorithm in  \citep{feldman2011unified} to achieve an approximation ratio of   $\frac{1}{e}-o(1)$. Let $A^{P02}$ denote the output from this algorithm. Note that $A^{P02}$  is not necessarily a feasible solution to $\textbf{P.0.1}$. This is because there may exist some $i\in [m]\setminus T$ such that $|A^{P02}\cap V_i|=0$. Let $T'\subseteq [m]\setminus T$ denote the indexes of those groups, i.e., $T'=\{i\in [m]\setminus T\mid |A^{P02}\cap V_i|=0\}$. Fortunately, by the definition of $T$, we have that for each $i\in [m]\setminus T$, we have $|V_i|\geq 2$. Hence, for each $i\in T'$, where $T'$ is a subset of $[m]\setminus T$,  $|V_i|\geq 2$. We pick two arbitrary items, say $x_i$ and $y_i$, from each group $i\in T'$, and build two candidate solutions as follows:
\[A^1=  A^{P02}\cup(\cup_{i\in T'} \{x_i\})\cup (\cup_{i\in T} V_i); A^2=  A^{P02} \cup(\cup_{i\in T'} \{y_i\})\cup (\cup_{i\in T} V_i).\]
Finally, we choose the better solution between $A^1$ and $A^2$ as the final solution $A^{\textsf{final}}$, that is, $f(A^{\textsf{final}})=\max\{f(A^1), f(A^2)\}$. We next show that
\begin{eqnarray}
\label{eq:kaison1}
\mathbb{E}[f(A^{\textsf{final}})]\geq \frac{1/e-o(1)}{2}f(OPT),
\end{eqnarray}
where the randomness is from $A^{P02}$.

Because $\cup_{i\in T'} \{x_i\}$ and $\cup_{i\in T'} \{y_i\}$ are disjoint and $f: 2^V\rightarrow \mathbb{R}_+$ is submodular, Lemma 1 in \citep{tang2021pointwise} implies that $f(A^{\textsf{final}})\geq f( A^{P02} \cup (\cup_{i\in T} V_i))/2$. Hence, $\mathbb{E}[f(A^{\textsf{final}})]\geq \mathbb{E}[f( A^{P02} \cup (\cup_{i\in T} V_i))]/2$. It follows that to prove (\ref{eq:kaison1}), it suffices to show that
\begin{eqnarray}
\label{eq:july}
\mathbb{E}[f( A^{P02} \cup (\cup_{i\in T} V_i))]\geq  (\frac{1}{e}-o(1))f(OPT).
\end{eqnarray}

The rest of the proof is devoted to proving (\ref{eq:july}). Let $S^{P02}$ denote the optimal solution to $\textbf{P.0.2}$. It follows that
\begin{eqnarray}
\mathbb{E}[f'(A^{P02})]\geq (\frac{1}{e}-o(1))f'(S^{P02})\geq (\frac{1}{e}-o(1))f'(S^{P01}),
\end{eqnarray}
where the second inequality is due to $\textbf{P.0.2}$ is a relaxation of $\textbf{P.0.1}$. By the definition of $f': 2^{V\setminus \cup_{i\in T} V_i}\rightarrow \mathbb{R}_+$, we further have
\begin{eqnarray}
&&\mathbb{E}[f(A^{P02}\cup(\cup_{i\in T} V_i))]\geq (\frac{1}{e}-o(1))f(S^{P02}\cup(\cup_{i\in T} V_i))\\
&&\geq (\frac{1}{e}-o(1))f(S^{P01}\cup(\cup_{i\in T} V_i)) \geq (\frac{1}{e}-o(1))f(OPT),
\end{eqnarray}
where the last inequality is due to (\ref{eq:hot}). This finishes the proof of (\ref{eq:july}).

\subsection{Solving the case when $\min_{i\in [m]}|OPT_i|\leq 1$}
 \label{sec:extended-speical-app}
 Now we are ready to discuss the case when $\min_{i\in [m]}|OPT_i|> 1$ does not hold.  Observe that if this condition does not hold, then $\min_{i\in [m]}|OPT_i|= 0$ or $1$. We next develop a  $\frac{1}{e}-o(1)$-approximation algorithm and a $\frac{1/e-o(1)}{2}$-approximation algorithm for these two cases, respectively. Although we do not know $\min_{i\in[m]}|OPT_i|$ initially, we can guess its value and solve the problem for each guess. Finally, the best solution is returned as the final output.

\subsubsection{$\min_{i\in [m]}|OPT_i|= 0$} For the case when $\min_{i\in [m]}|OPT_i|= 0$, i.e., the optimal solution selects zero items from some group, $OPT$ selects at most $\alpha$ items from any group because of the group equality constraint. Hence, $OPT$ is a feasible solution to the following optimization problem:
 \begin{center}
\framebox[0.6\textwidth][c]{
\enspace
\begin{minipage}[t]{0.6\textwidth}
\small
$\textbf{P.2.3}$
$\max f(S)$ \\
\textbf{subject to:}
$|S \cap V_i| \leq \alpha, \forall i \in[m]$ and
$|S|\leq c$.
\end{minipage}
}
\end{center}
\vspace{0.1in}

Note that $\textbf{P.2.3}$ is a submodular maximization problem subject to a matroid constraint. We can apply the algorithm in \citep{feldman2011unified} to find a $\frac{1}{e}-o(1)$-approximation solution, say $A^{\textsf{final}}$, for $\textbf{P.2.3}$. Because $\textbf{P.2}$ is a relaxation of  $\textbf{P.2.3}$, $A^{\textsf{final}}$ is feasible to  $\textbf{P.2}$. Meanwhile, because $OPT$ is a feasible solution to $\textbf{P.2.3}$ and $A^{\textsf{final}}$ is a $\frac{1}{e}-o(1)$-approximation solution for $\textbf{P.2.3}$, $A^{\textsf{final}}$ is a $\frac{1}{e}-o(1)$-approximation solution for $\textbf{P.2}$.

\subsubsection{$\min_{i\in [m]}|OPT_i|= 1$}
We next discuss the case when $\min_{i\in [m]}|OPT_i|= 1$. We  examine two subcases depending the value of $\alpha$. If $\alpha=0$, i.e., $OPT$ selects the same number of items from each group, then we have $|OPT_i|= 1$ for each $i\in[m]$ by the assumption that $\min_{i\in [m]}|OPT_i|= 1$. This indicates that $OPT$ must select all items from those groups whose size is one. Let $T=\{i\in[m]\mid |V_i|=1\}$ denote the set of the indexes of those groups whose size is one. Because $OPT$ select all items from $\cup_{i\in T} V_i$, it is safe to add $\cup_{i\in T} V_i$ to our solution in advance, leading to an optimization problem listed in $\textbf{P.2.4}$. The objective of  $\textbf{P.2.4}$ is $f'(\cdot)=f(\cdot\cup (\cup_{i\in T} V_i))$, which is a submodular function. It follows that $OPT\setminus \cup_{i\in T} V_i$ must be a feasible solution of the following problem.

 \begin{center}
\framebox[0.8\textwidth][c]{
\enspace
\begin{minipage}[t]{0.8\textwidth}
\small
$\textbf{P.2.4}$
$\max f'(S)$ \\
\textbf{subject to:}
$S\subseteq V\setminus \cup_{i\in T} V_i$ and
$|S \cap V_i| \leq 1, \forall i \in [m]\setminus T$ and
$|S| \leq c - |T|$.
\end{minipage}
}
\end{center}
\vspace{0.1in}

Note that the constraint $|S| \leq c - |T|$ can be omitted without affecting the global cardinality constraint. This is because in this case, the optimal solution $OPT$ selects exactly one item from each group, resulting in $|OPT|=m$. As $OPT$ is a feasible solution, it must satisfy the global cardinality constraint, which indicates that the global size constraint $c$ must be greater than or equal to $m$, i.e., $c\geq m$. Hence, any solution that selects at most one item from each group will also satisfy the global size constraint. Therefore, ensuring that $|S \cap V_i| \leq 1, \forall i \in [m]$, is sufficient to meet the global size constraint.

Because $f': 2^{V\setminus \cup_{i\in T} V_i}\rightarrow \mathbb{R}_+$ is a submodular function, $\textbf{P.2.4}$ is a classic submodular maximization problem subject to a matroid constraint. We apply the algorithm in  \citep{feldman2011unified} to achieve an approximation ratio of   $\frac{1}{e}-o(1)$. Let $A^{P24}$ denote the output from this algorithm. Note that $A^{P24}$  is not necessarily a feasible solution to $\textbf{P.2.4}$. This is because there may exist some $i\in [m]\setminus T$ such that $|A^{P24}\cap V_i|=0$, which violates the group equality constraint. Let $T'\subseteq [m]\setminus T$ denote the indexes of those groups, i.e., $T'=\{i\in [m]\setminus T\mid |A^{P24}\cap V_i|=0\}$. Fortunately, because $T'$ is a subset of $[m]\setminus T$,  we have $|V_i|\geq 2$ for all $i\in T'$ by the definition of $T$. We pick two arbitrary items, say $x_i$ and $y_i$, from each group $i\in T'$, and build two candidate solutions as follows:
\[A^1=  A^{P24}\cup(\cup_{i\in T'} \{x_i\})\cup (\cup_{i\in T} V_i); A^2=  A^{P24} \cup(\cup_{i\in T'} \{y_i\})\cup (\cup_{i\in T} V_i).\]

The better solution between $A^1$ and $A^2$ is returned as the final solution $A^{\textsf{final}}$, that is, $f(A^{\textsf{final}})=\max\{f(A^1), f(A^2)\}$. Following the same argument used to prove (\ref{eq:kaison1}), we have
\begin{eqnarray}
\label{eq:kaison2}
\mathbb{E}[f(A^{\textsf{final}})]\geq \frac{1/e-o(1)}{2}f(OPT).
\end{eqnarray}

We next discuss the case when $\alpha>0$. Given the optimal solution $OPT$, we pick an arbitrary item $e_i$ from each  $OPT_i$ and let $OPT^a=\cup_{i\in [m]} \{e_i\}$. Let $OPT^b=OPT\setminus OPT^a$. Because $f$ is submodular, we have
\begin{eqnarray}\label{eq:isr}
f(OPT^a)+f(OPT^b)\geq f(OPT).
 \end{eqnarray}

 To obtain an approximation  of $OPT$, it is sufficient to find an approximation of $OPT^a$ and $OPT^b$, respectively, then return the better one as the final output. It is easy to verify that $OPT^b$ is a feasible solution to $\textbf{P.2.3}$, hence, we can find a $\frac{1}{e}-o(1)$-approximation solution (labeled as $A^b$) by solving $\textbf{P.2.3}$. I.e., \begin{eqnarray}\label{eq:leila}
 \mathbb{E}[f(A^b)]\geq(\frac{1}{e}-o(1))f(OPT^b).
 \end{eqnarray}

 Meanwhile, $A^b$ is a feasible solution of our original problem. We next find an approximation of $OPT^a$. By the construction of $OPT^a$, it is easy to verify that $OPT^a$ is  a feasible solution to the following problem.

 \begin{center}
\framebox[0.6\textwidth][c]{
\enspace
\begin{minipage}[t]{0.6\textwidth}
\small
$\textbf{P.2.5}$
$\max f(S)$ \\
\textbf{subject to:}
$|S \cap V_i| \leq 1, \forall i \in[m]$ and
$|S|\leq c$.
\end{minipage}
}
\end{center}
\vspace{0.1in}

Because $\textbf{P.2.5}$ is a submodular maximization problem subject to a matroid constraint, we can find a $\frac{1}{e}-o(1)$-approximation solution (labeled as $A^a$) by solving $\textbf{P.2.5}$.  I.e.,
\begin{eqnarray}\label{eq:leila2}
\mathbb{E}[f(A^a)]\geq(\frac{1}{e}-o(1))f(OPT^a).
 \end{eqnarray}

 Meanwhile, $A^a$ must be a feasible solution of our original problem by the assumption that $\alpha>0$. Finally, we choose a better solution between $A^a$ and $A^b$ to achieve an approximation ratio of $\frac{\frac{1}{e}-o(1)}{2}$, i.e.,
\begin{eqnarray}
\mathbb{E}[\max\{f(A^a), f(A^b)\}]&&\geq \max\{\mathbb{E}[f(A^a)], \mathbb{E}[f(A^b)]\}]\\
&&\geq \frac{\mathbb{E}[f(A^a)]+ \mathbb{E}[f(A^b)]}{2}\\
&& \geq \frac{\frac{1}{e}-o(1)}{2}(f(OPT^a)+f(OPT^b))\\
&& \geq \frac{\frac{1}{e}-o(1)}{2} f(OPT),
\end{eqnarray}
where the third inequality is because of (\ref{eq:leila}) and (\ref{eq:leila2}); and the last inequality is due to (\ref{eq:isr}).

\subsection{Enhanced results for monotone case}
\label{sec:extended-monotone-app}
We next show that if the utility function $f$ is monotone, then we can achieve a $1/2$-approximation ratio. Suppose we know the value of  $z'=\min_{i\in[m]}|OPT\cap V_i|$ (if not, we can enumerate all $n$ possibilities of $\min_{i\in[m]}|OPT\cap V_i|$ and return the best solution as the final output), then solving our original problem $\textbf{P.2}$ is reduced to solving the following problem.
 \begin{center}
\framebox[0.8\textwidth][c]{
\enspace
\begin{minipage}[t]{0.8\textwidth}
\small
$\textbf{P.2.6}$
$\max f(S)$
\textbf{subject to:}
for each $i\in[m]$, $z' \leq |S\cap V_i|\leq z'+\alpha$ and
 $|S|\leq c$.
\end{minipage}
}
\end{center}
\vspace{0.1in}

To solve $\textbf{P.2.6}$, we  introduce another  problem $\textbf{P.2.7}$ as follows:
 \begin{center}
\framebox[0.9\textwidth][c]{
\enspace
\begin{minipage}[t]{0.9\textwidth}
\small
$\textbf{P.2.7}$
$\max f(S)$
\textbf{subject to:}
for each $i\in[m]$, $|S\cap V_i|\leq z'+\alpha$ and
$\sum_{i\in[m]}\max\{z', |S\cap V_i|\} \leq c$.
\end{minipage}
}
\end{center}
\vspace{0.1in}

Because $\textbf{P.2.7}$ is a relaxation of $\textbf{P.2.6}$, we have
\begin{eqnarray}
\label{eq:xing}
f(S^{P27})\geq f(S^{P26}),
\end{eqnarray} where $f(S^{P27})$ and $f(S^{P26})$ are optimal solutions of $\textbf{P.2.6}$ and $\textbf{P.2.7}$ respectively. As discussed earlier, if $f$ is monotone and submodular, then $\textbf{P.2.7}$ is a monotone submodular maximization problem subject to a matroid constraint. There exists a $(1-1/e)$-approximation algorithm for this problem. Let $A$ denote the output of this algorithm, we have $f(A)\geq (1-1/e)f(S^{P27})$. (\ref{eq:xing}) implies that $f(A)\geq (1-1/e)f(S^{P27})\geq  (1-1/e) f(S^{P26})$. If $A$ is a feasible solution of our original problem, then $A$ is returned as the final output. Otherwise, because $A$ is a feasible solution of $\textbf{P.2.7}$, then there must exist some $i\in[m]$ such that $|A\cap V_i|< z'$. In this case, we simply add $z'-|A\cap V_i|$ number of additional items from $V_i$ to $A$ for each $i\in[m]$ with $|S\cap V_i|< z'$ to obtain the final solution. It is easy to verify that this  solution satisfies both group equality and global cardinality constraints. Meanwhile, its utility is at least $f(A)$ because $f$ is a monotone function.

\end{APPENDICES}
% References here (outcomment the appropriate case)

% CASE 1: BiBTeX used to constantly update the references
%   (while the paper is being written).
%\bibliographystyle{ijocv081} % outcomment this and next line in Case 1
%\bibliography{<your bib file(s)>} % if more than one, comma separated

% CASE 2: BiBTeX used to generate mypaper.bbl (to be further fine tuned)
%\input{mypaper.bbl} % outcomment this line in Case 2

\end{document}